%% file: main.tex
\documentclass[lettersize,journal]{IEEEtran}
\usepackage{amsmath,amsfonts}
\usepackage{algorithmic}

\usepackage{array}
\usepackage[caption=false,font=normalsize,labelfont=sf,textfont=sf]{subfig}
\usepackage{textcomp}
\usepackage{stfloats}
\usepackage{url}
\usepackage{verbatim}
\usepackage{graphicx}
\usepackage{cite}
\usepackage{makecell}
\usepackage{amsmath,amssymb,amsfonts}
\usepackage{dsfont}

\usepackage{graphicx}
\usepackage{textcomp}
\usepackage[table,xcdraw]{xcolor}
\usepackage{array,multirow}
\usepackage[font=small]{caption}
\usepackage{balance}  

\usepackage{color, colortbl}
\definecolor{Gray}{gray}{0.8}
\definecolor{lGray}{gray}{0.9}
\usepackage{multicol}

\usepackage{enumerate}

\usepackage{enumitem}

\usepackage[normalem]{ulem}
\useunder{\uline}{\ul}{}

\usepackage[linesnumbered,algoruled,boxed,lined,noend]{algorithm2e}

\SetCommentSty{mycommfont}

\usepackage{comment}
\usepackage{url}
\newtheorem{problem}{Problem}

\newtheorem{definition}{Definition}

\newtheorem{lemma}{Lemma}

\newcommand{\etitle}[1]{\vspace{0.6ex}\noindent{\underline{\em #1}}}
\usepackage[normalem]{ulem}
\useunder{\uline}{\ul}{}
\usepackage{tikz}

\usepackage{pifont}

\newcommand{\eat}[1]{}

\newcommand{\rev}[1]{{\color{black} #1}\normalcolor}

\newcommand{\paul}[1]{{\color{black} #1}\normalcolor}
\newcommand{\paulJournal}[1]{{\color{black} #1}\normalcolor}

\newcommand{\dt}[1]{{\color{black} #1}\normalcolor}

\begin{document}

\title{$k$-Graph: A Graph Embedding for Interpretable Time Series Clustering}

\author{Paul Boniol, Donato Tiano, Angela Bonifati, Themis Palpanas}

\markboth{Journal of \LaTeX\ Class Files,~Vol.~14, No.~8, August~2021}%
{Shell \MakeLowercase{\textit{et al.}}: A Sample Article Using IEEEtran.cls for IEEE Journals}


\maketitle

\begin{abstract}
Time series clustering poses a significant challenge with diverse applications across domains. A prominent drawback of existing solutions lies in their limited interpretability, often confined to presenting users with centroids. In addressing this gap, our work presents $k$-Graph, an unsupervised method explicitly crafted to augment interpretability in time series clustering. Leveraging a graph representation of time series subsequences, $k$-Graph constructs multiple graph representations based on different subsequence lengths. This feature accommodates variable-length time series without requiring users to predetermine subsequence lengths. Our experimental results reveal that $k$-Graph outperforms current state-of-the-art time series clustering algorithms in accuracy, while providing users with meaningful explanations and interpretations of the clustering outcomes.
\end{abstract}


\input{Introduction.tex}

\input{Backprel.tex}
\input{Solution.tex}
\input{Experiments.tex}

\input{Conclusion.tex}

\section{acknowledgements}
Supported by EU Horizon projects AI4Europe (101070000), TwinODIS (101160009), ARMADA (101168951), DataGEMS (101188416), RECITALS (101168490).

\bibliographystyle{IEEEtran}
\bibliography{aaai22}

\begin{IEEEbiography}[{\includegraphics[width=1in,height=1.25in,clip,keepaspectratio]{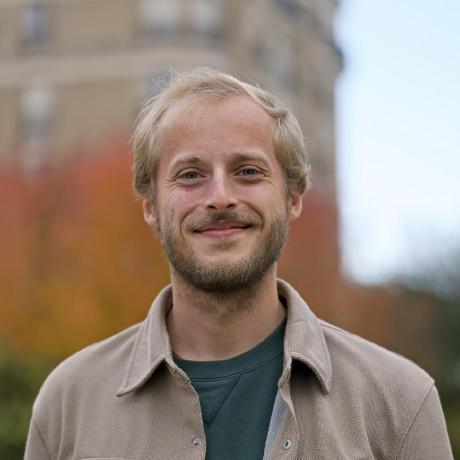}}]{Paul Boniol}
is a Researcher at Inria, in the VALDA project-team. Previously, he worked at ENS Paris-Saclay,
Université Paris Cité, EDF Research, and Ecole Polytechnique.
His research interests lie between data management, machine learning, and
time-series analysis. His Ph.D. focused on subsequence
anomaly detection and time-series classification, and won several PhD awards,
including the Paul Caseau Prize, supported by the Academy of
Sciences of France. His work has been published in the top data
management and data mining venues.
\end{IEEEbiography}

\vspace{-.65cm}

\begin{IEEEbiography}[{\includegraphics[width=1in,height=1.25in,clip,keepaspectratio]{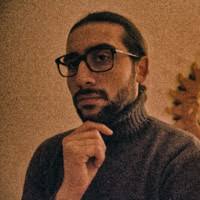}}]{Donato Tiano}is a Postdoctoral Researcher at the University of Modena and Reggio Emilia, part of the SoftLab group. He has worked at the University of British Columbia and the CNRS LIRIS lab at Lyon 1 University. His research focuses on interpretable solutions in time series analysis and text mining. Donato's work is published in leading data management and mining venues, where he also serves on the program committee.
\end{IEEEbiography}

\vspace{-.65cm}

\begin{IEEEbiography}[{\includegraphics[width=1in,height=1.25in,clip,keepaspectratio]{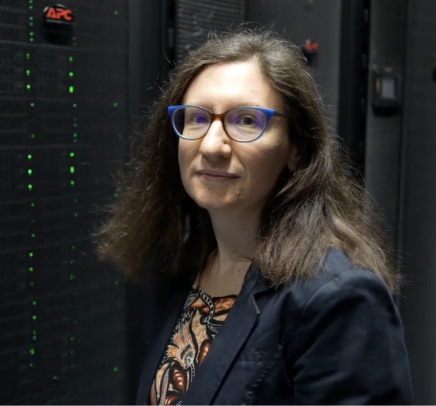}}]{Angela Bonifati}
is a Senior member of the French IUF and a Distinguished Professor in Computer Science at Lyon 1 University (France), affiliated with the CNRS Liris research lab. She is the Head of the Database group in the same lab. Since 2020, she is also an Adjunct Professor at the University of Waterloo (Canada) in the Data Systems Group.  She has co-authored more than 200 publications in data management, including five Best Paper Awards, two books and an invited ACM Sigmod Record paper. She is the recipient of the prestigious IEEE TCDE Impact Award 2023 and a co-recipient of an ACM Research Highlights Award 2023. She is the Program Chair of ICDE 2025, the General Chair of VLDB 2026 and an Associate Editor for the Proceedings of VLDB and for several other journals (VLDB Journal, IEEE TKDE, ACM TODS, etc.).
\end{IEEEbiography}

\vspace{-.65cm}

\begin{IEEEbiography}[{\includegraphics[width=1in,height=1.25in,clip,keepaspectratio]{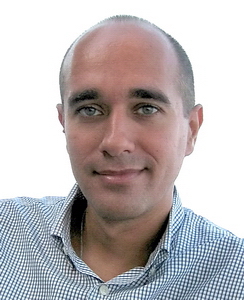}}]{Themis Palpanas}
is an elected Senior Member of the French
University Insitute (IUF), and Distinguished Professor of computer
science at Universite Paris Cite (France). He has authored 14 patents,
received 3 best paper awards and the IBM SUR award, has been
Program Chair for VLDB 2025 and IEEE BigData 2023, General Chair
for VLDB 2013, and has served Editor in Chief for BDR. He has been
working in the fields of Data Series Management and Analytics for
more than 15 years, and has developed several of the state of the art
techniques. 
\end{IEEEbiography}

\vfill

\end{document}

%% file: Introduction.tex
\section{Introduction}
\label{sec:intro}
Massive collections of time series are becoming a reality in virtually every scientific and social domain~\cite{Palpanas2019,MAHDAVINEJAD2018161}. 
Thus, there is a significant need for multiple relevant applications for proposing methods that can efficiently analyze them~\cite{DBLP:journals/dagstuhl-reports/BagnallCPZ19}. 
Examples of fields that involve data series are finance, environmental sciences, astrophysics, neuroscience, engineering, multimedia, etc.~\cite{fulfillingtheneed,DBLP:journals/dagstuhl-reports/BagnallCPZ19}. 
Once collected, the major analysis tasks on  time series include pattern matching (or similarity search)~\cite{c19-isip-Palpanas-isaxfamily, DBLP:journals/tkde/PengFP21, seanet}, classification~\cite{DBLP:journals/datamine/YeK11,DBLP:conf/cikm/SchaferL17, DBLP:journals/datamine/TanPW20, inceptionTime}, clustering~\cite{DBLP:conf/sdm/UlanovaBK15,DBLP:journals/sigmod/PaparrizosG16,DBLP:conf/ijcai/Li0Z19}, anomaly detection~\cite{benchref,Series2GraphPaper,DBLP:conf/edbt/Gao0B20,normajournal,DBLP:journals/pvldb/BoniolPPF21} and motif discovery~\cite{DBLP:journals/tkde/ZhuMK21}.

Time series clustering poses a pivotal and complex challenge in data science, garnering substantial attention with many proposed algorithms in recent years. Traditional approaches hinge on distance measures, exemplified by $k$-Means clustering utilizing Euclidean or Dynamic Time Warping (DTW) distances, and the widely adopted $k$-Shape algorithm~\cite{DBLP:journals/sigmod/PaparrizosG16} serves as a prominent baseline method.
However, recent studies underscore the efficacy of feature-based methods, where clustering operates on extracted time series features, showcasing robust performance in accuracy and execution time~\cite{bonifati2022time2feat}. 
Despite their success, these methods grapple with noteworthy limitations: (i) susceptibility to noisy time series, which can compromise clustering performance, and (ii) a need for more interpretability in most solutions, hindering a comprehensive understanding of the outcomes. The last limitation is a challenging problem, especially when the important features to discover are typical subsequences that could be a grouping criterion for a given cluster. Therefore, identifying such subsequences and providing an interpretable representation of the clustering partitions to the user is essential to enhance the usage and maximize the understanding and trust of unsupervised clustering in time series applications. Unfortunately, none of the existing methods allow a straightforward solution for this problem.

\begin{figure}[tb]
 \centering
\includegraphics[width=\linewidth]{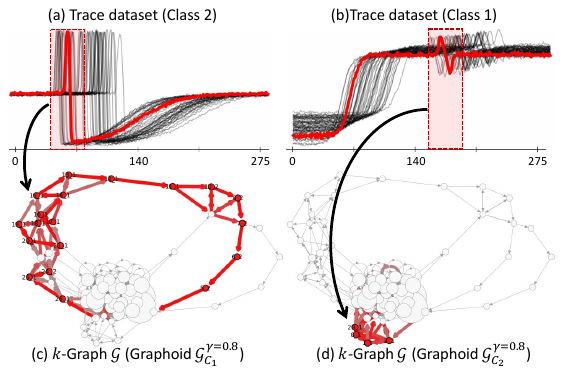}
 \caption{$k$-Graph resulting graph $\mathcal{G}$ when applied on the Trace dataset~\cite{Dau2018TheUT}}
 \vspace{-0.3cm}
 \label{fig:intro}
\end{figure}

This paper introduces $k$-Graph, a novel graph-based \rev{univariate time series } clustering method that aims to overcome the shortcomings of existing approaches and the problems mentioned above. \rev{To the best of our knowledge, $k$-Graph is the first approach proposing a graph-based representation of the time series for the purpose of clustering. Overall, }
$k$-Graph is based on a four-step process: The first step involves \textbf{Graph Embedding} where several graphs are computed leveraging an adapted Series2Graph algorithm~\cite{Series2GraphPaper}. Each graph encapsulates groups of similar subsequences within a dataset, with nodes representing these groups and edges carrying weights based on sequence occurrences. Moving forward, in the \textbf{Graph Clustering} phase, features are extracted for time series using the graph nodes and edges. The $k$-Means algorithm is then applied for clustering based on these features. Next, spectral clustering is deployed in \textbf{Consensus Clustering} to establish consensus across the multiple partitions obtained in the previous step. This step yields the final labels assigned by $k$-Graph. Lastly, the most relevant graph is selected in \textbf{Interpretability Computation}, and graphoids are computed to enhance interpretability \paulJournal{(examples of graphoids for the Trace dataset~\cite{Dau2018TheUT} are depicted in Figure~\ref{fig:intro})}. This step contributes to a more profound understanding of the clustering outcomes.

Our contributions are as follows:
\begin{enumerate}
    \item We provide a new problem formulation for graph embedding for time series clustering (Section~\ref{sec:Probdef}). We formalize the concept of interpretability regarding clustering with graph representation of time series and how to measure it.
    \item We propose a novel graph embedding method for \rev{univariate time series } clustering. The latter can be used on time series of variable length and provides through the graph an interpretable interface for the user to dive into the time series datasets and extract meaningful patterns that compose the different clusters (Section~\ref{sec:proposed}).
    \item We demonstrate that our proposed approach is at least equalling and outperforming under certain scenarios, the current state-of-the-art clustering methods for time series, and being of the same order of magnitude regarding execution time (Section~\ref{sec:exp}).
    \item In addition to being as accurate and as expensive to compute, we demonstrate through multiple examples on real world datasets the interpretability power of $k$-Graph, and the usefulness of the graph for knowledge discovery tasks (Section~\ref{sec:exp}).
    \item We provide an open-source implementation of our approach (\url{https://github.com/boniolp/kGraph}). 
\end{enumerate}

We conclude by discussing the implications of our work and exploring future directions to enhance the accuracy, execution time, and interpretability of our proposed method (Section~\ref{sec:conclusions}).

%% file: Backprel.tex
\section{Background and Related Work}
\label{sec:background}

We first introduce
notations useful for the rest of the paper (Section~\ref{sec:notation}). Then, we review
existing time-series clustering methods and discuss their limitations related to interpretability and user interaction (Section~\ref{sec:clus_methods}). We then motivate and introduce the usage of Graph embedding for time series clustering (Section~\ref{sec:graphforts}). We finally properly define the problem tackled in this paper (Section~\ref{sec:Probdef}).

\subsection{Time Series and Graph Notation}
\label{sec:notation}

\textbf{Time Series: }A \rev{univariate } time series $T \in \mathbb{R}^n $ is a sequence of real-valued numbers $T_i\in\mathbb{R}$ $[T_1,T_2,...,T_n]$, where $n=|T|$ is the length of $T$, and $T_i$ is the $i^{th}$ point of $T$. \rev{In the rest of this paper , we refer to univariate time series as {\it time series}. } We are typically interested in local regions of the time series, known as subsequences. A subsequence $T_{i,\ell} \in \mathbb{R}^\ell$ of a time series $T$ is a continuous subset of the values of $T$ of length $\ell$ starting at position $i$. Formally, $T_{i,\ell} = [T_i, T_{i+1},...,T_{i+\ell-1}]$.
A dataset $\mathcal{D}$ is a set of time series (possibly of different lengths). 
We define the size of $\mathcal{D}$ as $|\mathcal{D}|$.

\noindent \textbf{Graph: }
We introduce some basic definitions for graphs, which we will use in this paper.
We define a Node Set $\mathcal{N}$ as a set of unique integers.
Given a Node Set $\mathcal{N}$, an Edge Set $\mathcal{E}$ is then a set composed of tuples $(x_i,x_j)$, where $x_i,x_j \in \mathcal{N}$. 
Given a Node Set $\mathcal{N}$, an Edge Set $\mathcal{E}$ (pairs of nodes in $\mathcal{N}$), a Graph $\mathcal{G}$ is an ordered pair $\mathcal{G}=(\mathcal{N},\mathcal{E})$.
A directed graph $\mathcal{G}$ is an ordered pair $\mathcal{G}=(\mathcal{N},\mathcal{E})$ where $\mathcal{N}$ is a Node Set, and $\mathcal{E}$ is an ordered Edge Set.
In the rest of this paper, we will only use directed graphs, denoted as $\mathcal{G}$.

\subsection{Time Series Clustering}
\label{sec:clus_methods}
Time series clustering plays a pivotal role in uncovering meaningful patterns within temporal datasets, where the primary goal is to group similar time series for insightful analysis. The specific challenge addressed by partitional time series clustering involves partitioning a dataset of $n$ time series, denoted as $D$, into $k$ clusters $(C = \{C_{1}, C_{2}, ..., C_{k}\})$, where the number of clusters, $k$, is predetermined. Although the choice of $k$ can be determined using wrapper methods, it is assumed to be fixed in advance in many experiments. 

\subsubsection{Raw-Based Approaches}

Clustering algorithms for time series can either operate directly on raw data or use transformations to derive features before clustering. Algorithms specifically designed for time series often prefer using raw data~\cite{liao2005clustering} for several reasons. First, \textbf{Preservation of Information}: raw time series preserve intricate details of temporal patterns, ensuring information fidelity. Second, \textbf{Temporal Dependencies}: they capture dynamic patterns and temporal interdependencies that might be lost during feature extraction. Finally, \textbf{Data Exploration}: direct analysis of raw time series fosters the discovery of unexpected patterns and trends, supporting an exploratory, discovery-driven approach.

In the raw-based approach, the $k$-Means algorithm is commonly used to identify patterns and similarities within temporal data. Each time series is treated as a multidimensional vector, with data points representing observations over time. The goal is to partition the time series into $k$ clusters, optimizing the assignment of series to cluster centers.

$k$-Shape~\cite{10.1145/2949741.2949758} is a well-known algorithm for time series clustering in this approach. It starts by randomly selecting initial cluster centers and uses a specialized distance metric to measure dissimilarity based on shapes, addressing temporal dynamics. The algorithm iteratively assigns each time series to the cluster with the smallest shape-based distance and updates cluster centers based on the mean shape. This process continues until convergence, resulting in clusters that highlight shape similarities and provide insights into temporal dynamics.

\rev{A recent study~\cite{10.14778/3611540.3611622} has compared $k$-Shape with a large amount of clustering baselines and has demonstrated that none of the recent baselines outperform significantly $k$-Shape. }
\rev{Finally, several raw-basd appraoch are using the Dynamic Time Warping (DTW), a well-known algorithm for measuring the similarity between two time series by aligning them through time axis warping~\cite{10.1145/2783258.2783286}. This process minimizes the distance between sequences, accommodating shifts and stretches in the time dimension. Although DTW is highly accurate, it is computationally intensive, often necessitating optimizations for improved efficiency. SOMTimeS~\cite{10.1007/s10618-023-00979-9} is an algorithm that addresses this challenge. It is a self-organizing map (SOM) designed for clustering time series data, utilizing DTW as a distance measure to effectively align and compare time series. SOMTimeS enhances computational efficiency by implementing a pruning strategy that significantly reduces the number of DTW calculations required during the training.} 

\dt{However, it is important to note that raw-based approaches may face challenges when dealing with noise in raw time series, potentially obscuring meaningful patterns. Additionally, the direct clustering of raw time series might result in clusters lacking clear distinctions or meaningful insights, making extracting valuable information from the data challenging.}

\subsubsection{Feature-Based Approaches}
Adopting a feature-based \cite{wang2006characteristic} approach offers several advantages that address the challenges associated with raw-based methods: 
\textbf{Dimensionality Reduction}, feature extraction often reduces dimensionality. Indeed, lower-dimensional feature representations enhance computational efficiency and reduce the risk of the curse of dimensionality; \textbf{Enhanced Discrimination}, feature selection can be tailored to emphasize specific characteristics crucial for discrimination, enhancing the ability of clustering algorithms to distinguish subtle differences between time series, leading to more accurate clustering; and \textbf{Interpretability}, clusters derived from features often yield more interpretable results than those directly from raw time series. Interpretability is crucial for extracting insights, and features-based approaches provide clearer explanations for grouping.

\rev{TS3C~\cite{8960542} is an example of feature-based clustering approach. The latter segments the time series and extracts a set of statistical features to represent them. The features are then used to cluster time series. }
FeatTS~\cite{tiano2021featts} emerges as a feature-based algorithm tailored for univariate time series clustering. It leverages the TSFresh library and employs Principal Features Analysis to extract salient features from time series data. 
Building upon this foundation, the authors introduce Time2Feat (T2F)~\cite{bonifati2022time2feat}, an algorithm designed explicitly for multivariate time series clustering. T2F distinguishes itself by adopting two distinct feature extraction approaches named intra and inter-signal feature extraction. 
Furthermore, the authors enhance the feature selection process by introducing a grid search method for selecting optimal features for clustering multivariate time series.
FeatTS and Time2Feat are two approaches that accommodate unsupervised and semi-supervised clustering scenarios. Notably, Time2Feat stands out as the current state-of-the-art solution for multivariate time series clustering, showcasing the advancements in feature extraction and selection methodologies.

\dt{Nevertheless, feature-based approaches may suffer from the information loss induced by the transformation of the original time series (i.e., sequential patterns) into features. 
Handling high-dimensional feature spaces poses a challenge, especially when dealing with numerous derived features. This complexity can compromise the interpretability of the algorithm, making it difficult for users to examine all the extracted features.
Additionally, the sensitivity of feature engineering poses a challenge, where selecting inappropriate features or applying unsuitable transformations may affect clustering quality.}

\subsubsection{Deep Learning Approaches}

In recent years, deep learning strategies~\cite{alqahtani2021deep} have emerged to address the challenges of time series clustering, demonstrating strong performance. These approaches leverage deep learning's ability to directly interface with time series data, marking a significant shift from traditional methods. A key advantage is the \textbf{Hierarchy of Abstraction}, where deep learning architectures capture complex relationships within temporal data, revealing intricate patterns essential for effective clustering. Additionally, these methods exhibit \textbf{High Efficacy}, surpassing traditional techniques with superior accuracy and efficiency.

The Deep Auto-Encoder (DAE)~\cite{tian2014learning} is a popular solution, serving as an unsupervised model for representation learning. It transforms raw input data into new space representations, extracting valuable features through encoding. The DAE architecture, characterized by seven fully connected layers, effectively harnesses learned features through an internal layer. These features are subsequently input into a clustering loss function, minimizing the distance between data points and their respective assigned cluster centers.

Another approach is Deep Temporal Clustering (DTC)~\cite{DTCAlgorithm}, which uses the DAE for feature representation and clustering. The DTC's clustering layer optimizes the Kullback-Leibler (KL) divergence objective, aligning with a self-training target distribution. The encoding process influences clustering performance, with learned representations fed into the $k$-Means algorithm for final clustering.

\dt{In conclusion, despite the high efficacy and the ability to uncover complex patterns, deep learning approaches encounter fundamental challenges related to the interpretability of their decision-making processes. The inherent complexity of these machine learning models often results in a lack of transparency in understanding obtained results. Furthermore, these approaches may struggle to integrate domain-specific knowledge seamlessly, presenting obstacles to guiding the clustering process based on expert insights.}

\subsection{Graph Embedding of Time Series}
\label{sec:graphforts}

We argue that to address the challenges of maintaining information preservation while maximizing interpretability, a viable solution is to represent time series as a (suitably constructed) graph. 
Constructing such a graph involves processing subsequences of the time series dataset. 
These subsequences represent various types of patterns and their temporal succession. This approach furnishes clustering methods with substantial information, contributing to the sustenance of high accuracy. Additionally, this approach facilitates user-friendly navigation through the time series.

Various methods have been proposed to convert time series into graphs for specific analytical tasks to overcome the challenges mentioned earlier. 
Series2Graph embeds univariate time series into a directed graph~\cite{Series2GraphPaper, DistributedS2G}, primarily employed for anomaly detection. Similarly, approaches like Time2Graph~\cite{9477138} utilize graph representations of time series to address time series classification.
The advantages of such time series graph representations are threefold. 
First, these graph representations are easily interpretable by any user. Second, constructed directly from subsequences of the time series, they preserve essential information. Lastly, a unified embedding can significantly reduce execution time, as evidenced in anomaly detection scenarios~\cite{DistributedS2G}.

However, prior works on graph embedding for time series were often proposed either under supervised settings~\cite{9477138}, simplifying the graph construction task, or in an unsupervised manner but for continuous time series~\cite{Series2GraphPaper}.
For the specific task of time series clustering, a graph embedding of time series necessitates the unsupervised construction of a single, comprehensive graph for an entire dataset, encompassing multiple continuous time series. While a straightforward solution would be building one graph per time series, this approach diminishes the interpretability advantage of having a single, concise graph.

\subsection{Problem Formulation}
\label{sec:Probdef}

\begin{figure}[tb]
 \centering
\includegraphics[width=\linewidth]{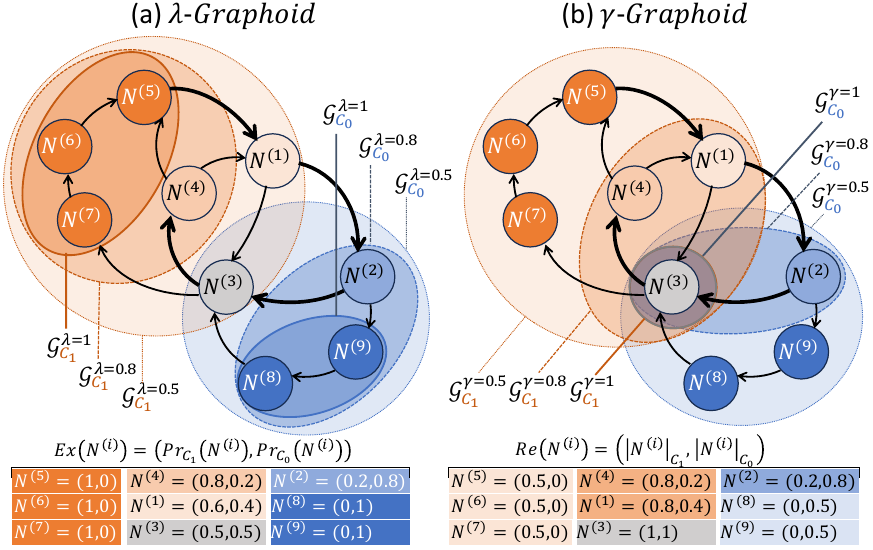}

 \caption{$\lambda$-Graphoids and $\gamma$-Graphoids for different $\lambda$ and $\gamma$.}
 \label{fig:def}
\vspace{-0.3cm}
\end{figure}

We propose a novel 
approach to time series clustering by employing a graph embedding. The essence of this methodology lies in transforming a time series dataset into a sequence of abstract states corresponding to different subsequences within the dataset. These states are represented as nodes, denoted by $\mathcal{N}$, in a directed graph, $\mathcal{G}=(\mathcal{N},\mathcal{E})$. The edges, $\mathcal{E}$, encode the frequency with which one state occurs after another~\cite{Series2GraphPaper}. We define this graph as follows:

\begin{definition}[Graph Embedding]   
Let a time series dataset be defined as $\mathcal{D} = \{T_1,T_2,...,T_n\}$. Let $\mathcal{G}=(\mathcal{N},\mathcal{E})$ be a directed graph. 
$\mathcal{G}$ is the graph embedding of $\mathcal{D}$ if there exists a function $\mathcal{M}$ such that for any $T \in \mathcal{D}$, $\mathcal{M}(T) =  \langle N^{(1)},N^{(2)},...,N^{(m)} \rangle $, and  $\forall i \in [1,m], N^{(i)} \in \mathcal{N}$, and $\forall i \in [1,m-1], (N^{(i)},N^{(i+1)}) \in \mathcal{E}$. 
\label{defGraph}  
\end{definition}

The rest of the section considers $\mathcal{M}(T)$ as a subgraph.
For the sake of interpretability, it is imperative that different sections of the graph distinctly capture the similarity between time series. Nodes, representing similar patterns from various time series, and edges, denoting possible transitions between these patterns, play a pivotal role in this context. Consequently, a comparable time series is found to be placed within the same region of the graph.
As a result, in a given dataset $\mathcal{D} = \{T_0, T_1, ..., T_n\}$, a designated cluster of time series, denoted as $C_i \subset \mathcal{D}$, corresponds to a discernible subgraph $\mathcal{G}_{C_i} \subset \mathcal{G}$. This subgraph is formally called a ``\textit{Graphoid}."

\begin{definition}[$Graphoid$]   
Let $\mathcal{D}$ be a time series dataset and $C_i \subset \mathcal{D}$ a given cluster such as $C_i = \{T_1,T_2,...,T_{k'}\}$. Let $\mathcal{G}$ be the graph embedding of $\mathcal{D}$ resulting from a function $\mathcal{M}$. We define the {\it Graphoid} of $C_i$ as:
{\small
\[
\mathcal{G}_{C_i} = \bigcup_{T \in C_i} \mathcal{M}(T)
\]
}
\label{defGraphoid}  
\end{definition}

In Definition~\ref{defGraphoid}, the {\it Graphoid} of a given cluster contains all the nodes and edges that at least one time series of that cluster crossed. Therefore, a node of $\mathcal{G}$ may belong to multiple graphoids. As a consequence, no distinction is made between nodes that contain subsequences of all time series of one cluster (i.e., {\bf representativity}) or time series subsequences of one cluster only (i.e., {\bf exclusivity}) and nodes that are contained in all clusters. We thus introduce the two following definitions:

\begin{definition}[Node representativity]   
Let $\mathcal{D}$ a time series dataset and $C = \{C_1,C_2,...,C_k\}$ a clustering partition. Let $\mathcal{G}=(\mathcal{N},\mathcal{E})$ be the graph embedding of $\mathcal{D}$ resulting from a function $\mathcal{M}$. We define the {\it Representativity} of node $N \in \mathcal{N}$ as $Re(N) = (|N|_{C_1}, ..., |N|_{C_k}$ with $|N|_{C_i}$ defined as:
{\small
\[
|N|_{C_i} = \frac{1}{{|C_i|}}\sum_{T \in C_i} 1_{[N \in \mathcal{M}(T)]}
\]
}
\label{defnodeRep}  
\end{definition}

\begin{definition}[Node Exclusivity]   
Let $\mathcal{D}$ a time series dataset and $C = \{C_1,C_2,...,C_k\}$ a clustering partition. Let $\mathcal{G}=(\mathcal{N},\mathcal{E})$ be the graph embedding of $\mathcal{D}$ resulting from a function $\mathcal{M}$. We define the {\it Exclusivity} of node $N \in \mathcal{N}$ as $Ex(N) = (Pr_{C_1}(N), ..., Pr_{C_k}(N))$ with $Pr_{C_i}(N)$ defined as:
{\small
\[
Pr_{C_i}(N) = \frac{|C_i||N|_{C_i}}{\sum_{T \in \mathcal{D}} 1_{[N \in \mathcal{M}(T)]}}
\]
}
\label{defnodeEx}  
\end{definition}

In other words, the {\bf representativity} of a node is the number of time series of a given cluster that crossed the node divided by the total number of time series within that cluster. 
The {\bf exclusivity} of a node is the number of time series of a given cluster that crossed the node divided by the total number of time series that crossed that same node. The same definitions can be used for edges. 
Based on the above definitions, we can restrict the definition of a {\it Graphoid} based on exclusivity and representativity. 
We thus introduce $\lambda$-$Graphoid$ and $\gamma$-$Graphoid$ defined as follows:

\begin{definition}[$\lambda$-$Graphoid$]   
For a given dataset $\mathcal{D}$, $\mathcal{G}$ the graph embedding of $\mathcal{D}$, and a given cluster $C_i$. The $\lambda$-$Graphoid$ of $C_i$ is defined as $\mathcal{G}^{\lambda}_{C_i} = (\mathcal{N}^{\lambda}_{C_i},\mathcal{E}^{\lambda}_{C_i})$ such as $\forall N \in \mathcal{N}^{\lambda}_{C_i}, \forall E \in \mathcal{E}^{\lambda}_{C_i}, Pr_{C_i}(N) \geq \lambda$ and $Pr_{C_i}(E) \geq \lambda$.
\label{deflambdaGraph}  
\end{definition}

\begin{definition}[$\gamma$-$Graphoid$]   
For a given dataset $\mathcal{D}$, $\mathcal{G}$ the graph embedding of $\mathcal{D}$, and a given cluster $C_i$. The $\gamma$-$Graphoid$ of $C_i$ is defined as $\mathcal{G}^{\gamma}_{C_i} = (\mathcal{N}^{\gamma}_{C_i},\mathcal{E}^{\gamma}_{C_i})$ such as $\forall N \in \mathcal{N}^{\gamma}_{C_i}, \forall E \in \mathcal{E}^{\gamma}_{C_i}, |N|_{C_i} \geq \gamma$ and $|E|_{C_i} \geq \gamma$.
\label{defgammaGraph}  
\end{definition}

The concepts introduced above can be better illustrated using Figure~\ref{fig:def}. The $\lambda$-$Graphoid$ and $\gamma$-$Graphoid$ are influenced by the chosen values of $\lambda$ and $\gamma$. As demonstrated in Figure~\ref{fig:def}, higher values of $\lambda$ and $\gamma$ lead to more restrictive graphoids. In the illustration, nodes $N^{(5)},N^{(6)},N^{(7)}$ are exclusively crossed by time series from cluster $C_1$, highlighting unique patterns specific to this cluster. On the other hand, node $N^{(3)}$ is traversed by all time series of $C_1$ and is considered the most representative pattern for this cluster. However, it also represents a common pattern for $C_0$.

This scenario emphasizes the trade-off between representativity and exclusivity. While $N^{(3)}$ provides a comprehensive representation of $C_1$, it lacks exclusivity to this cluster. In contrast, $N^{(5)}, N^{(6)}, N^{(7)}$ offer exclusive patterns for $C_1$ but might not be present in all time series of this cluster, limiting the interpretability. Thus, finding an optimal balance between higher values of $\lambda$ and $\gamma$ is crucial for maximizing the interpretability of a clustering partition through graph embedding.
Based on the above definitions, we can state the following:

\begin{lemma}
    For a given clustering partition $C = \{C_1,C_2,...,C_k\}$, if $\lambda \leq k$, then $\bigcup_{C_i \in C} \mathcal{G}^{\lambda}_{C_i} = \mathcal{G}$.
    if $\lambda > 0.5$, then $\bigcap_{C_i \in C} \mathcal{G}^{\lambda}_{C_i} = \emptyset$.
    \label{Lemma1} 
\end{lemma} 

\begin{lemma}
    For a given clustering partition $C = \{C_1,C_2,...,C_k\}$, if $\forall C_i \in C, \mathcal{G}^{\lambda=1}_{C_i} = \mathcal{G}^{\gamma=\frac{1}{|C_i|}}_{C_i}$, then $\bigcap_{C_i \in C} \mathcal{G}_{C_i} = \emptyset$
    \label{Lemma2} 
\end{lemma} 

Lemma~\ref{Lemma2} corresponds to the perfect partition, in which each graph region exclusively represents one cluster. 
However, we do not need to have all the nodes of a {\it Graphoid} exclusively representing only one cluster. 
It is sufficient to have one node for each cluster with $|N|_{C_i} = Pr_{C_i}(N) = 1$.  
Therefore, the problem we want to solve is the following.

\begin{problem}[Time Series Graph Clustering]
	Given a dataset $\mathcal{D}$, automatically construct the graph $\mathcal{G}(\mathcal{N},\mathcal{E})$, and compute a clustering partition $C = \{C_1,...,C_k\}$ of $\mathcal{D}$, such that:  
{\small
 \[
 \forall C_i \in C, |\mathcal{G}^{\lambda=1}_{C_i}  \cap \mathcal{G}^{\gamma=1}_{C_i} |> 0
 \]
 }
 \label{problem}
\end{problem}

As this problem is impossible to solve in some use cases, the objective is to find the largest possible values of $\lambda$ and $\gamma$, such that the condition in Problem~\ref{problem} holds and the values of $\lambda$ and $\gamma$ indicate the quality of the clustering interpretability. Table~\ref{SymbolTable} summarizes the symbols used in this paper.

\begin{table}[tb]
\centering
\begin{tabular}{c|c}
\hline
{\bf Symbol} & {\bf Description} \\
\hline
$T$									& a time series (of length $|T|$) \\
$\ell$ 								& subsequence length\\
$\mathcal{D}$ 						& a dataset of time series\\
$C_i$							    & a cluster of a clustering partition $C$\\
$\mathcal{L}$						& labels generated by clustering\\
$k$							        & number of clusters\\
$\mathcal{N},\mathcal{E}$			& set of nodes and edges\\
$d$			& degrees of a set of nodes\\
$\mathcal{G}$			            & directed graph\\
$\mathcal{M}$ 						& function that transform $T$ into $\mathcal{G}$ \\
$\mathcal{G}_{\ell}$ 				& graph embedding built with length $\ell$ \\
$F_{\mathcal{G},\ell}$ 						& feature matrix for graph $\mathcal{G}_\ell$ \\
$M_C$ 								& consensus matrix \\
$\mathcal{G}_{C_i}$,$\mathcal{G}^{\lambda}_{C_i},\mathcal{G}^{\gamma}_{C_i}$ 		& graphoid, $\lambda$-graphoid and $\gamma$-graphoid of $C_i$ \\
$|N|_{C_i}$, $Pr_{C_i}(N)$						& representativity, exclusivity of node $N$ for $C_i$\\
\hline
\end{tabular}
\caption{Table of symbols}
\label{SymbolTable}
\end{table}

%% file: Solution.tex
\section{Proposed Approach}
\label{sec:proposed}

\begin{figure*}[tb]
 \centering
\includegraphics[width=0.95\linewidth]{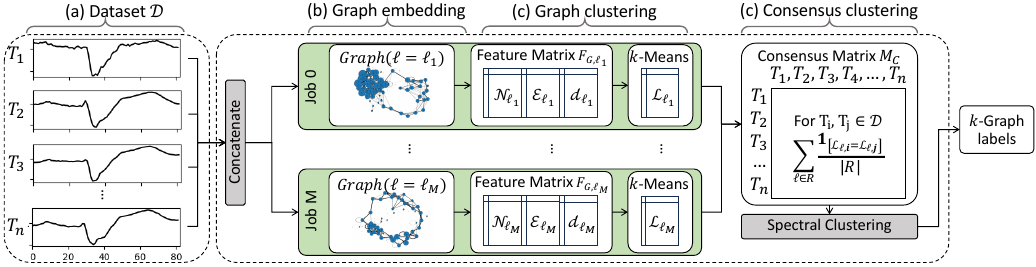}
\vspace{-0.2cm}
 \caption{$k$-Graph pipeline.}
 \label{fig:pipeline}
\vspace{-0.3cm}
\end{figure*}

In this section, we describe $k$-Graph, an approach for time series clustering, as a solution to tackle the problem described in the previous section. 
For a given dataset $\mathcal{D}$, the overall $k$-Graph process is divided into three main steps as follows:

\begin{enumerate}[noitemsep, topsep=1pt, parsep=1pt, partopsep=1pt, leftmargin=0.5cm]
\item {\bf Graph Embedding} (Section~\ref{sol:graphembedding}): for $M$ different subsequence lengths, we compute $M$ graphs.
For a given subsequence length $\ell$, The set of nodes represent groups of similar subsequences of length $\ell$ within the dataset $\mathcal{D}$. The edges have weights corresponding to the number of times one subsequence of a given node has been followed by a subsequence of the other node.

\item {\bf Graph Clustering} (Section~\ref{sol:graphclustering}): For each graph, we extract a set of features for all time series of the dataset $\mathcal{D}$. These features correspond to the nodes \paul{and edges}
that the time series crossed. Then, we use these features to cluster the time series using $k$-Means \paul{for scalability reasons}.

\item {\bf Consensus Clustering} (Section~\ref{sol:consensusclustering}): At this point we have $M$ clustering partitions (i.e., one per graph). \paul{We build a consensus matrix $M_C$. We then cluster this matrix using spectral clustering in the objective of grouping time series that are highly connected (i.e., grouped in the same cluster in most of the $M$ partitions)}. 
The output of this clustering step is the labels provided by $k$-Graph.

\item {\bf Interpretability Computation} (Section~\ref{sol:interpretability}): after obtaining the clustering partition, we select the most relevant graph (among the $M$ graphs), and we compute the \paul{interpretable} graphoids.
\end{enumerate}

\IncMargin{0.5em}
\begin{algorithm}[tb]
{\small
    \caption{\textbf{$k$-Graph}}\label{alg:kgraph}
    \SetKwInOut{Input}{input}
    \SetKwInOut{Output}{output}
    \Input{Dataset $\mathcal{D}$, Number of clusters $k$, Number of length $M$ (default is 30), Maximum length rate $rml$ (default is 0.4), Sample rate $smpl$ (default is 10)}
    \Output{Clustering labels $\mathcal{L}$}
    \BlankLine
    $all$-$\mathcal{L}$ $\leftarrow$ $[]$ \;
    $R$ $\leftarrow$ $M$ random lengths in $[5,(min_{T \in \mathcal{D}} |T|)*rml]$ \;
    \ForEach{$\ell \in R$ }
    {
        $\mathcal{G}_\ell \leftarrow Graph(\mathcal{D},\ell,smpl)$ \tcp{Section~\ref{sol:graphembedding}}
        $F_{\mathcal{G},\ell} \leftarrow extractFeature(\mathcal{D},\mathcal{G}_\ell)$ \tcp{Section~\ref{sol:graphclustering}}
        $\mathcal{L}_{\ell} \leftarrow k$-$Means(F_{\mathcal{G},\ell},k)$ \tcp{Section~\ref{sol:graphclustering}}
        add $\mathcal{L}_{\ell}$ in $all$-$\mathcal{L}$\;
    }
    $M_C$ $\leftarrow$ $Consensus(all$-$\mathcal{L})$ \tcp{Section~\ref{sol:consensusclustering}}
    $\mathcal{L}$ $\leftarrow$ $SpectralClustering(M_c,k)$ \tcp{Section~\ref{sol:consensusclustering}}

    return $\mathcal{L}$

}
\end{algorithm}
\DecMargin{0.5em}

Figure~\ref{fig:pipeline} illustrates the different steps of $k$-Graph, detailed in Algorithm~\ref{alg:kgraph}.
Below, we describe each step in detail.

\subsection{Graph Embedding}
\label{sol:graphembedding}

As described in Section~\ref{sec:Probdef}, the general objective is to build one graph representing the time series dataset on which we can perform clustering. 
\paul{For this step, a solution could be to use the Series2Graph algorithm~\cite{Series2GraphPaper}. 
Nevertheless, Series2Graph presents three limitations. 
First, it accepts a continuous time series as input, rather than a dataset, that may contain several series of different lengths. 
Second, the user must specify a subsequence length $\ell$. 
This becomes problematic in interpretable clustering, as determining an optimal length in advance is challenging. 
Third, the embedding employed by Series2Graph does not scale to large dataset sizes, which would be a significant drawback in our case. 
In $k$-Graph, we propose the following graph embedding procedure. 

For a dataset $\mathcal{D}$, we build $M$ different graphs $\mathcal{G}_\ell=(\mathcal{N}_\ell,\mathcal{E}_\ell)$ for $M$ different subsequence length $\ell$. Algorithm~\ref{alg:GraphEmbedind} outlines the procedure for constructing a single graph for a given dataset and subsequence length.
The procedure is the following:

\begin{enumerate}[noitemsep, topsep=1pt, parsep=1pt, partopsep=1pt, leftmargin=0.5cm]
    \item {\bf Subsequence Embedding}: For each time series $T \in \mathcal{D}$, we collect all the subsequences of a given length $\ell$ into an array called $Proj(T,\lambda)$. We then concatenate all the computed $Proj(T,\lambda)$ into $Proj$ for all the time series in the dataset (Line~\ref{concatenation}).
    We then sample $Proj$ (user-defined parameter $smpl$) and keep only a limited number of subsequences stored in $Proj_{smpl}$. We use the latter to train a Principal Component Analysis (PCA) (Line~\ref{pcatrain}). We then use the trained PCA and a rotation step to project all the subsequences into a two-dimensional space that only preserves the shapes of the subsequences. The result is denoted as $SProj$. We denote the PCA and rotation steps $Reduce(Proj,pca)$, where $pca$ is the trained PCA.
    
    \item {\bf Node Creation}: Create a node for each of the densest parts of the above two-dimensional space~\cite{Series2GraphPaper}. 
    In practice, we perform a radial scan of $SProj_{smpl}$.
    For each radius, we collect the intersection with the trajectories of $SProj_{smpl}$, and we apply kernel density estimation on the intersected points: each local maximum of the density estimation curve is assigned to a node.
    These nodes can be seen as a summarization of all the major length patterns $\ell$ that occurred in $\mathcal{D}$. For this step, we only consider the sampled collection of subsequences $SProj_{smpl}$. We denote this step $NodeCr(SProj_{smpl})$ (Line~\ref{nodec}).

    \item {\bf Edge Creation}: Retrieve all transitions between pairs of subsequences represented by two different nodes: each transition corresponds to a pair of subsequences, where one occurs immediately after the other in a time series $T$ of the dataset $\mathcal{D}$. We represent transitions with an edge between the corresponding nodes. We note this step $EdgeCr(SProj,\mathcal{N}_\ell)$ (Line~\ref{nodec}). $\mathcal{N}_\ell$ is the node set extracted in the above step.
\end{enumerate}
}
\IncMargin{0.5em}
\begin{algorithm}[tb]
{\small
    \caption{\textbf{Graph Embedding $Graph(\mathcal{D},\ell)$}}\label{alg:GraphEmbedind}
    \SetKwInOut{Input}{input}
    \SetKwInOut{Output}{output}
    \Input{Dataset $\mathcal{D}$, input length $\ell$, sample $smpl$}
    \Output{graph $\mathcal{G}_\ell$}
    \BlankLine
    \tcp{Concatenante subsequences}
    $Proj \leftarrow []$\;
    \ForEach{$T \in \mathcal{D}$ }
    {
        add $Proj(T,\ell)$ in $Proj$\;\label{concatenation}
    }
    \tcp{Reduce to a two-dimensional space}
    $Proj_{smpl}$ $\leftarrow$ randomly select $\lfloor\frac{|Proj|}{smpl}\rfloor$ elements in $Proj$\;
    $pca$ $\leftarrow$ $PCA.fit(Proj_{smpl})$\;\label{pcatrain}
    $SProj$ $\leftarrow$ $Reduce(Proj,pca)$\;

    \tcp{Extract the nodes and edges}
    $\mathcal{N}_\ell, \mathcal{E}_\ell \leftarrow NodeCr(SProj_{smpl}), EdgeCr(SProj,\mathcal{N}_\ell)$\;\label{nodec}

    return $\mathcal{G}_\ell = (\mathcal{N}_\ell,\mathcal{E}_\ell)$

} 
\end{algorithm}
\DecMargin{0.5em}

\rev{Despite the linear aspect of the PCA (in the subsequence embedding step), we observed in practice that using the first three components was sufficient to embed enough shape-based information~\cite{Series2GraphPaper}. Furthermore, PCA is renowned for its ability to identify and emphasize principal patterns while diminishing the impact of noise~\cite{Soltysik2015ImprovingTU,Lei2014SparsistencyAA}. The latter guarantees a graph-based representation less affected by the presence of noise in the time series subsequences.}

As mentioned above, the graph embedding requires a parameter $smpl$ that controls the number of subsequences used to train the PCA and to create the nodes. 
We set by default the parameter $smpl=10$ and evaluate its influence in Section~\ref{sec:exp}. 
Moreover, we generate the graph embedding using Algorithm~\ref{alg:GraphEmbedind} for $M$ different lengths randomly selected from a predefined interval. 
We denote the set of randomly selected lengths as $R$, where the user can specify the number of lengths by setting $M = |R|$. 
Our observations show that accuracy and execution time tend to improve with an increase in $M$. However, the accuracy reaches a plateau after $M=30$, as demonstrated in Section~\ref{sec:exp}.

Since the subsequence embedding step necessitates subsequences of at least 5 points, the length interval can be constrained to $[5, \min_{T \in \mathcal{D}} |T|]$. 
However, empirical observations indicate that this interval can be further optimized. We define this interval as $[5, ( \min_{T \in \mathcal{D}} |T|) \times rml]$, where $rml$ (rate maximum length) is a user-defined parameter with a default value of 0.4 
(evaluated in Section~\ref{sec:exp}).

Figure \ref{fig:pipeline}, we show the graphs for $3$ different lengths ($\ell$ equal to 10, 30, and 50) for the TwoLeadECG dataset of the UCR-Archive~\cite{Dau2018TheUT}. 
The variation in the graph's topology emphasizes the significance of subsequence length. 
Building multiple graphs ensures that $k$-Graph does not rely on a predetermined subsequence length, thus avoiding situations where an erroneous or unsuitable choice of the single subsequence length leads to a suboptimal graph, potentially failing to capture critical data patterns.

\subsection{Graph Clustering}
\label{sol:graphclustering}

At this stage, we have $M$ distinct graphs corresponding to $M$ different subsequence lengths. 
To perform a clustering partition of the dataset $\mathcal{D}$ using these graphs, we opt for a feature-based approach for execution time and interpretability. 
Our goal is to maximize information extraction from the graph. For this purpose, we extract \paulJournal{three} types of features for each time series: (i) node-based, (ii) edge-based features, \paulJournal{and (iii) degree-based features.} Specifically, \paulJournal{for two first types of features}, we quantify how many times a given time series intersects each node and edge in the graph. \paulJournal{For the third type, we measure the degree for each node in the subgraph corresponding to each time series. For example, the same node can have a high degree for one time series (corresponding to a specific subgraph in $\mathcal{G}_\ell$) and a low degree for another time series (corresponding to another subgraph).}
Formally, for a given dataset $\mathcal{D}$ and its graph embedding $\mathcal{G}_\ell=(\mathcal{N}_\ell,\mathcal{E}_\ell)$ computed using a subsequence length $\ell$, we define for a time series $T \in \mathcal{D}$, $\mathcal{M}_\ell(T) = \langle N^{(1)},N^{(2)},...,N^{(m)} \rangle$ the path of $T$ in $\mathcal{G}_\ell$, \paulJournal{ and $\mathcal{G}_\ell(T)$ the corresponding subgraph.} 
Thus, we define the feature matrix $F_{\mathcal{G},\ell} \in \mathbb{R}^{(|\mathcal{D}|,|\mathcal{N}_\ell|+|\mathcal{E}_\ell|)}$ as follows:

\paulJournal{
{\small
\begin{align*}
\begin{split}
\forall T_i \in \mathcal{D}, F_{\mathcal{G},\ell}[i] = [f_{T_i,N_1},...,f_{T_i,N_{|\mathcal{N}_\ell|}}, &f_{T_i,E_1},...,f_{T_i,E_{|\mathcal{E}_\ell|}}, \\
&f^d_{T_i,N_1},...,f^d_{T_i,N_{|\mathcal{N}_\ell|}}] \\
\end{split}
\end{align*}
}

With $\forall N,E \in \mathcal{N}_\ell,\mathcal{E}_\ell$:
{\small
\begin{align*}
\begin{split}
    f_{T_i,N} &= \sum_{N^{(j)} \in \mathcal{M}_\ell(T_i)} 1_{[N=N^{(j)}]} \\
    f_{T_i,E} &= \sum_{(N^{(j)},N^{(j+1)}) \in \mathcal{M}_\ell(T_i)} 1_{[E=(N^{(j)},N^{(j+1)})]}\\
    f^{d}_{T_i,N} &=  deg_{\mathcal{G}_\ell(T)}(N)*1_{[N=(N^{(j)},N^{(j+1)})]}
\end{split}
\end{align*}
}
}
\paulJournal{In the equation above, $deg_{\mathcal{G}_\ell(T)}(N)$ denotes the degree of the node $N$ within the subgraph $\mathcal{G}_\ell(T)$.}
As described above, $F_{\mathcal{G},\ell}$ is a sparse matrix with values greater than zero only if a time series crosses a node or an edge. 
However, as time series can be of different lengths, the number of nodes and edges crossed can vary significantly from one dataset instance to another. 
As a consequence, we normalize each row $F_{\mathcal{G},\ell}[i]$ by subtracting the its mean $\mu(F_{\mathcal{G},\ell}[i])$ and dividing by its standard deviation $\sigma(F_{\mathcal{G},\ell})$. 
We can then apply any clustering algorithm on $F_{\mathcal{G},\ell}$. 

\rev{We use $k$-Means with Euclidean distance to produce a partition $\mathcal{L}_\ell$ for a given graph $\mathcal{G}_\ell$. Normalization is applied only during the clustering step, while the unnormalized feature matrix is retained for computing graphoids. The choice of $k$-Means is motivated by its scalability for large datasets and its centroid-based approach, which aids interpretability. The centroids from graph clustering are crucial for computing graphoids, which are key to interpreting the clustering results of k-Graph (see Section~\ref{sol:interpretability}).}

\subsection{Consensus Clustering}
\label{sol:consensusclustering}

At this point, we have one clustering partition $\mathcal{L}_\ell$ per graph built in the graph embedding step ($M$ in total, for $\ell$ in the set of randomly selected lengths $R$). 
The objective is to compute a consensus from all these partitions. 
The problem of consensus clustering, or ensemble clustering, is a well-studied problem with several methods proposed in the literature~\cite{Strehl2003ClusterE,Fred2005CombiningMC}.

Following established practice~\cite{Strehl2003ClusterE}, we build a consensus matrix, which we employ to measure how many times two time series have been grouped in the same cluster for two graphs built with two different lengths. 
Formally, we define $M_C \in \mathbb{R}^{(|\mathcal{D}|,|\mathcal{D}|)}$ a matrix computed as follows:
\begin{equation}
\small{
    \forall T_i,T_j \in \mathcal{D}, M_C[i,j] = \frac{1}{|R|}\sum_{\ell \in R} 1_{[\mathcal{L}_{\ell,i} = \mathcal{L}_{\ell,j}]}
}
\end{equation}

$M_C$ can be seen as a similarity matrix about the clustering results obtained on each graph. More specifically, for two time series $T_i$ and $T_j$, if $M_C[i,j]$ is high, they have been associated in the same cluster for several subsequence lengths and can be grouped in the same cluster. On the contrary, if $M_C[i,j]$ is low, the two time series were usually grouped in different clusters regardless of the subsequence length. \paul{Therefore, the $M_C$ matrix can be seen as the adjacency matrix of a graph. 
In this graph, nodes are the time series of the dataset, and an edge exists if two time series have been clustered together in the same cluster (the weights of these edges are the number of times these two time series have been clustered together). 
As the objective is to find communities of highly connected nodes (i.e., time series that were grouped multiple times in the same cluster), we use spectral clustering (with $M_C$ used as a pre-computed similarity matrix)}. 
The output of the spectral clustering is the final labels $\mathcal{L}$ of $k$-Graph.

\subsection{Interpretability and Explainability}
\label{sol:interpretability}

\begin{figure}[tb]
 \centering
\includegraphics[width=\linewidth]{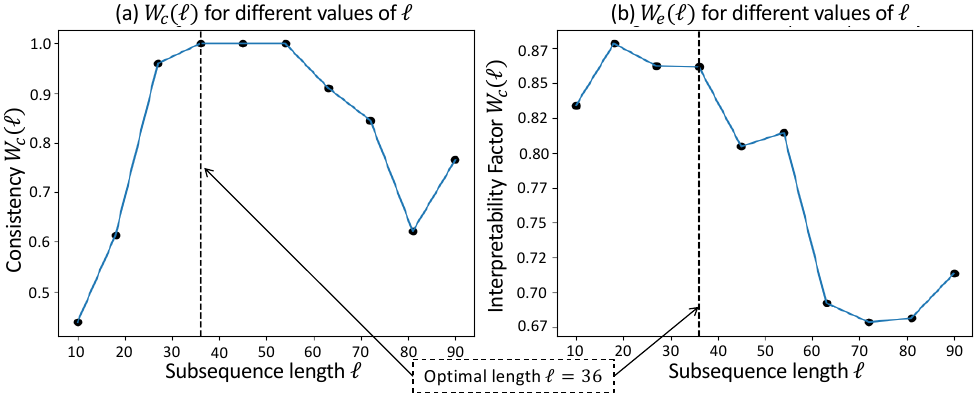}

 \caption{$W_c$ and $W_e$ for $k$-Graph (with $k=4$ and $M=10$) applied on the Trace dataset of the UCR-Archive.}
\vspace{-0.3cm}
 \label{fig:lengthselection}
\end{figure}

The previous section elucidated $k$-Graph's approach to the time series clustering problem. 
Now, we delve into the steps necessary for solving the interpretable clustering problem. 
As outlined in Section~\ref{sec:Probdef}, we formalized the interpretability problem through the computation of graphoids. 
The challenge lies in the assumption that there is a unique graph. 
However, $k$-Graph operates with M graphs to produce one dataset partition. 
We are presented with two options: (i) compute the graphoids for each graph, or (ii) select one graph that is most representative of the final labeling $\mathcal{L}$.

Considering option (i), we assume equal contributions from each graph to the labels $\mathcal{L}$. 
However, this assumption falters due to the consensus clustering step. 
For instance, among the M graphs, one-third might generate one type of partition, while the remaining two-thirds yield another. 
Consequently, the partition from the latter two-thirds would be selected as the final labels $\mathcal{L}$ of $k$-Graph. 
As a result, one-third of the graphs might be irrelevant for the final labels, yielding graphoids that do not correctly interpret the clustering.

Hence, we opt for option (ii), choosing one of the M graphs: the one that is most relevant to the clustering labels $\mathcal{L}$ produced by $k$-Graph. 
To make this selection, we employ two criteria: the consistency of $\mathcal{L}_\ell$ with $\mathcal{L}$, and the interpretability factor of $\mathcal{G}_\ell$ associated with $\mathcal{L}$.

\noindent{\bf [Consistency]} To gauge the consistency between the final labels $\mathcal{L}$ and those generated by each graph $\mathcal{G}_\ell$ (for $\ell \in R$), we utilize the Adjusted Rand Index (ARI). \rev{The latter is particularly suitable for unbalance datasets as it accounts for expected similarity due
to chance (See Section~\ref{sec:exp}). } Formally, the consistency is expressed as $W_c(\ell) = ARI(\mathcal{L},\mathcal{L}_\ell)$, measuring the agreement between the clustering labels produced by $k$-Graph and those corresponding to a specific subsequence length $\ell$. 
This index provides a quantitative similarity measure, with higher values indicating greater consistency.

For a practical illustration, Figure~\ref{fig:lengthselection}(a) displays the values of $W_c$ across different subsequence lengths ($\ell$) in the range $R=[10,28,27,36,45,54,63,72,81,90]$. 
In this example, $k$-Graph, configured with $k=4$ and $M=10$, is applied to the Trace dataset from the UCR-Archive. 
The graph generated with subsequence lengths between $27$ and $54$ demonstrates high consistency (ARI values above $0.9$) with the final labels $\mathcal{L}$, indicating that these lengths are more suitable for interpretation. 
Conversely, other lengths result in lower consistency, suggesting that labels produced under these conditions are less aligned with the final clustering labels of $k$-Graph. 
Therefore, for optimal interpretability, it is advisable to consider subsequence lengths within the range of $27$ to $54$.

\noindent{\bf [Interpretability Factor]} The second criterion focuses on selecting the most interpretable graphs in tandem with a given clustering partition. 
As elucidated in Section~\ref{sec:Probdef}, interpretability is achieved by ensuring that each cluster has precisely one node, characterized by $|N|_{C_i}=Pr_{C_i}(N) = 1$. 
To quantify interpretability, we calculate the interpretability factor as the maximum $Pr_{C_i}(N)$ across all nodes $N$ in the graph $\mathcal{G}_\ell$.
Formally, let $\mathcal{D}$ be a dataset, and $\mathcal{G}_\ell=(\mathcal{N}_\ell,\mathcal{E}_\ell)$ be its graph embedding computed with a subsequence length $\ell$. 
We denote $C={C_1,...,C_k}$ as the clustering partition associated with labels $\mathcal{L}$ obtained using $k$-Graph. The interpretability factor $W_e(\ell)$ of $\mathcal{G}_\ell$ is defined as:

\begin{equation}
\small{
    W_e(\ell) = \frac{1}{k} \sum_{C_i \in C} \max_{N \in \mathcal{N}_\ell} Pr_{C_i}(N)
}
\end{equation}

In practical terms, the $\mathcal{M}$ function in Definition~\ref{defnodeEx} is instantiated as follows: for a time series $T \in \mathcal{D}$, $\mathcal{M}_\ell(T) = \langle N^{(1)},N^{(2)},...,N^{(m)} \rangle$, representing the path of $T$ in $\mathcal{G}_\ell$. 
Figure~\ref{fig:lengthselection}(b) illustrates $W_e$ for the Trace dataset of the UCR-Archive, revealing that graphs computed with shorter lengths exhibit a higher interpretability factor.
To select the optimal length $\bar{\ell}$ and, consequently, the graph $\mathcal{G}_\ell$ that maximizes both $W_c$ and $W_e$, we employ the following criterion:

\begin{equation}
\small{
    \bar{\ell} = \underset{\ell \in R}{\operatorname{argmax}} \big[W_e(\ell).W_c(\ell)\big]
    }
\end{equation}

In Figure~\ref{fig:lengthselection}, the sole length that maximizes the product $W_e(\ell) \cdot W_c(\ell)$ is $\bar{\ell}=36$. Subsequently, the graph associated with this length is utilized to compute the graphoids.

To compute the graphoids, we use the feature matrix $F_{G,\bar{\ell}}$ (described in Section~\ref{sol:graphclustering}). In practice, we compute the representativity and the exclusivity of a node $N$ in $\mathcal{G}_\ell$. We can compute the two previous measures for all nodes in $\mathcal{G}_\ell$, and then select only those that satisfy the property of $\lambda$-Graphoid and $\gamma$-Graphoid for a desired value of $\lambda$ and $\gamma$. As one highly representative and exclusive node per cluster is enough, we automatically return one node $\bar{N}_{C_i}$ per cluster such that:

\begin{equation}
\small{
    \bar{N}_{C_i} = \underset{N \in \mathcal{N}_\ell}{\operatorname{argmax}} \big[|N|_{C_i}.Pr_{C_i}(N)\big]
}
\end{equation}

The node $\bar{N}_{C_i}$ is considered the most interpretable node for a given cluster $C_i$. Finally, for each node, we store the timestamps of the subsequences in $\mathcal{D}$ associated with it. Therefore, we compute the centroid of all the subsequences that compose $\bar{N}_{C_i}$. Figure~\ref{fig:interpretability} and Figure~\ref{fig:interpretability_ex} show the most interpretable nodes (and the corresponding centroids) identified by $k$-Graph for several example datasets.

\rev{In our approach, interpretability (i.e., graphoid for each cluster) is integrated during the training phase rather than applied post-hoc. By embedding interpretability into the clustering process, we ensure that the selected graph consistently represents the clustering logic during both training and subsequent predictions. This ensures reliable interpretability, even when incoming time series exhibit variability or contain multiple relevant subsequences.}

\subsection{Complexity Analysis}
\label{sol:complexity}

This section analyzes the computational complexity of different steps in $k$-Graph. 
For the Graph Embedding step, the sequence embedding complexity is bounded by the PCA (using a randomized SVD solver) with complexity $O(|\mathcal{D}|^2\cdot|T|^2)$. 
The node creation and edge creation steps are both $O(|\mathcal{D}|\cdot|T|)$ on average.
Following the graph embedding step, the complexity for feature matrix creation and $k$-Means clustering is $O(|\mathcal{D}|\cdot(|\mathcal{N}|+|\mathcal{E}|))$ and $O(k.|\mathcal{D}|\cdot(|\mathcal{N}|+|\mathcal{E}|))$ respectively. In practice, the number of nodes and edges is significantly smaller than the number of subsequences in the dataset, making this step negligible compared to the graph embedding step. 
Finally, building the consensus matrix requires $O(M\cdot|\mathcal{D}|^2)$. 
As the consensus matrix is already built, spectral clustering requires, in the worst case, $O(|\mathcal{D}|^2)$. 
Overall, since the first two steps are executed $M$ times (for $M$ different subsequence lengths), the $k$-Graph complexity is $O(M\cdot|\mathcal{D}|^2\cdot|T|^2)$. 
However, as the computations of the $M$ graphs are independent, they can be executed in parallel, significantly reducing execution time. 
Moreover, the $smpl$ parameter significantly reduces $|\mathcal{D}|$. 
Section~\ref{sec:exp} evaluates the influence of $M$ and $smpl$ on execution time.

%% file: Experiments.tex
\section{Experimental Evaluation}
\label{sec:exp}

\label{sec:setup}

\noindent{\bf [Setup]} We implemented our algorithms in Python 3.9. The evaluation was conducted on a server Intel(R) Xeon(R) Gold 6242R CPU 3.10GHz (80 CPUs), with 512GB RAM.
Our code is publicly available: \url{https://github.com/boniolp/kGraph}

\begin{figure*}[tb]
 \centering
\includegraphics[width=0.97\linewidth]{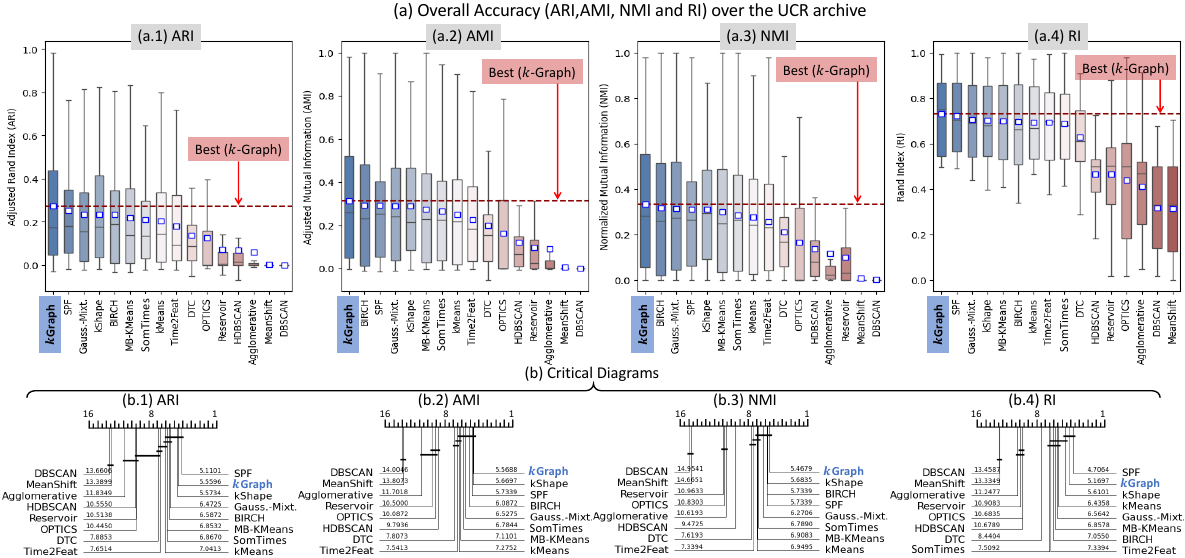}

 \caption{\rev{Experimental comparison of $k$-Graph versus the baselines on the UCR archive. In (a), the mean values are represented as a white square. The horizontal red dotted line represents the best mean.}}
 \vspace{-0.3cm}
 \label{fig:experiments_effectiveness_figures}

\end{figure*}

\label{sec:configuration_data_baseline}
\noindent{\bf [Datasets]} We conducted an experimental evaluation utilizing real datasets from the UCR-Archive~\cite{Dau2018TheUT} to assess the classification performance of various methods. While our proposed approach is versatile and can be evaluated on variable-length time series, the baseline methods do not share this flexibility. Therefore, out of the initial 128 datasets, we excluded 15 containing variable-length time series. Our comparative analysis involves assessing our proposed approach against the baselines on a subset of 113 real-time series datasets.
Table~\ref{tab:dataset_statistics} summarizes the characteristics of the adopted datasets, 
which represent several different scenarios.

\begin{table}[tb]
\centering
\scalebox{0.92}{
\begin{tabular}{r|r|r|r|r}
\hline
\multicolumn{1}{l}{Dataset Type} & \multicolumn{1}{l}{\# Dataset} & \multicolumn{1}{l}{Avg. \# TS} & \multicolumn{1}{l}{Avg. Length} & \multicolumn{1}{l}{Avg. Classes} \\
\hline
AUDIO & 1 & 2110.00 & 1024.00 & 39.00 \\
DEVICE & 11 & 2428.18 & 809.27 & 3.55 \\
ECG & 7 & 2313.71 & 513.29 & 14.14 \\
EOG & 2 & 724.00 & 1250.00 & 12.00 \\
EPG & 2 & 288.50 & 601.00 & 3.00 \\
H.DYN. & 3 & 312.00 & 2000.00 & 52.00 \\
IMAGE & 31 & 1713.81 & 338.45 & 10.61 \\
MOTION & 6 & 343.83 & 899.00 & 3.83 \\
OTHER & 1 & 204.00 & 201.00 & 18.00 \\
SENSOR & 14 & 2213.64 & 348.79 & 3.07 \\
SIMUL. & 9 & 1566.33 & 255.44 & 3.89 \\
SOUND & 11 & 1982.27 & 308.18 & 6.91 \\
SPECTRO & 12 & 448.17 & 980.75 & 3.42 \\
TRAFFIC & 1 & 365.00 & 24.00 & 2.00 \\
\hline
\end{tabular}
}
\caption{UCR-Archive Dataset category Statistics}
\label{tab:dataset_statistics}
\vspace{-0.3cm}
\end{table}

\noindent{\bf [Baselines]} 
In our experimental evaluation, we consider the 15 state-of-the-art (SOTA) algorithm from each category: 
\begin{itemize}[noitemsep, topsep=1pt, parsep=1pt, partopsep=1pt, leftmargin=0.5cm]

    \item \textbf{Traditional Clustering:} We include the $k$-Means algorithm, which uses the Euclidean distance as a similarity measure. \paulJournal{We also include eight additional traditional clustering approaches for completeness which are Mean-Shift~\cite{1000236} (M.Sh. in Table~\ref{tab:effectiveness_category}), Gaussian Mixture~\cite{10.5555/1953048.2078195} (G.M. in Table~\ref{tab:effectiveness_category}), BIRCH~\cite{10.1145/233269.233324}, MiniBatch-$k$-Means~\cite{10.5555/1953048.2078195}, OPTICS~\cite{10.1145/304182.304187}, HDBSCAN~\cite{10.1007/978-3-642-37456-2_14}, DBSCAN~\cite{10.5555/3001460.3001507} and Agglomerative Clustering with a average linkage~\cite{10.5555/1953048.2078195} (Aggl. in Table~\ref{tab:effectiveness_category}). For all these baselines, we use their corresponding scikit-learn~\cite{10.5555/1953048.2078195} implementation, and we select their default parameters.}
    
    \item \textbf{Raw-based Approaches:} We select the $k$-Shape algorithm~\cite{10.1145/2949741.2949758}. This algorithm identifies the most discriminative sub-shapes of the time series to form the final cluster. \rev{In addition, we adopt SomTimeS, an algorithm that uses a self-organizing map (SOM) framework to align and compare time series with Dynamic Time Warping (DTW) as the distance measure. It enhances computational efficiency through a pruning strategy that reduces the number of DTW calculations during training.}

    \item \paulJournal{\textbf{Symbolic Approaches:} We select the Symbolic Pattern Forest algorithm (SPF)~\cite{ijcai2019-406}. The approach checks if some randomly selected symbolic patterns exist in the time series to partition the dataset. This partition process is executed multiple times, and the ensemble combines the partitions to generate the final partition.}
    
    \item \textbf{Features-based Approaches:} We choose Time2Feat~\cite{bonifati2022time2feat} (T2F in Table~\ref{tab:effectiveness_category}). This approach extracts and selects the best features from the time series, providing a fully interpretable and effective solution.

    \item \textbf{Deep Learning Approaches:} We opt for two baselines. First, the Deep Temporal Clustering (DTC) algorithm~\cite{DTCAlgorithm}. DTC uses feature representation extracted from the Deep Auto-Encoder and the Kullback-Leibler (KL) divergence objective for self-labeling of the dataset. \paulJournal{The second baseline considered is the Reservoir Model Space~\cite{bianchi2020reservoir} (called Reservoir in the rest of the paper and Res. in Table~\ref{tab:effectiveness_category}). Reservoir Computing (RC) is a family of Recurrent Neural Network (RNN) models whose recurrent part is generated randomly and then kept fixed. The representation learned with Reservoir is then fed into a traditional clustering.}

\end{itemize}

\noindent{\bf [Metrics]} We compare $k$-Graph against state-of-the-art approaches using metrics that assess how well clusters align with predefined or expected data structures by comparing them with external information like labels.
We focus on four metrics: the Adjusted Rand Index (ARI)~\cite{steinley2004properties}, Adjusted Mutual Information (AMI), Normalized Mutual Information (NMI), and Rand Index (RI).

\rev{The Adjusted Rand Index (ARI), derived from the Rand Index (RI), measures the similarity between two clusterings by counting pairs of data points in the same or different clusters. Unlike the RI, the ARI accounts for expected similarity due to chance, which is crucial for accurate results, especially in imbalanced datasets. This correction provides a more reliable measure of clustering performance, with a scale from -0.5 to 1. A value of 1 indicates perfect agreement, 0 suggests chance-level agreement and negative values indicate disagreement. By considering class balance, the ARI offers a more robust assessment of clustering similarity compared to the RI}. To enhance evaluation robustness, we also use Adjusted Mutual Information (AMI) and Normalized Mutual Information (NMI). Mutual Information (MI) measures the mutual dependence between two clusterings, but cluster sizes and numbers can influence it. NMI normalizes MI to a scale from 0 (no mutual information) to 1 (perfect correlation). AMI further adjusts MI for chance, providing a normalized measure that is useful for comparing clustering results across datasets with varying cluster sizes.

\subsection{Accuracy Evaluation}

\begin{figure}[tb]
 \centering
\includegraphics[width=\linewidth]{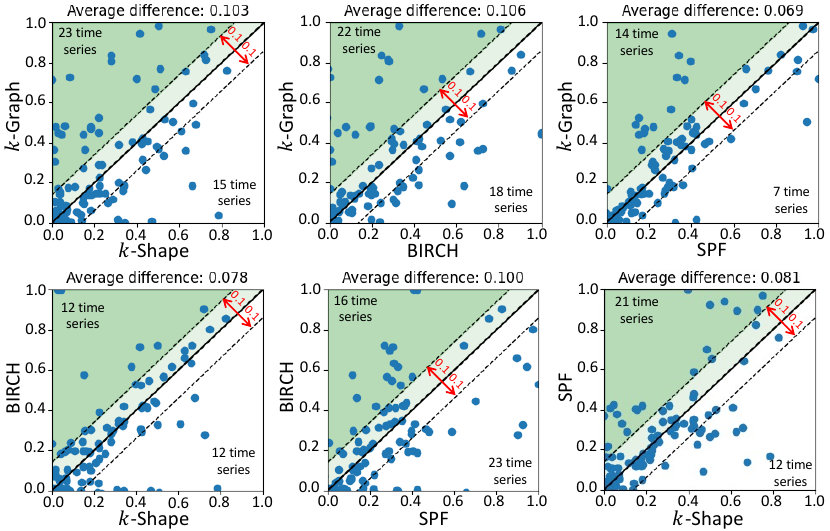}

 \caption{Pairwise comparisons (ARI) of the top performing methods, namely, $k$-Graph, $k$-Shape, SPF, and BIRCH.}
\vspace{-0.3cm}
 \label{fig:pairwise_comp}
\end{figure}

\etitle{Presentation} Table~\ref{tab:effectiveness_category} showcases the performance of $k$-Graph compared to state-of-the-art approaches, expressed in terms of Adjusted Rand Index (ARI). The datasets are categorized as introduced in Section~\ref{sec:configuration_data_baseline}.
\paulJournal{We also include the average ARI, NMI, AMI, and RI (for the entire UCR-Archive), as well as the number of wins and the average rank (for each measure).}
The results demonstrate $k$-Graph's superior performance across various dataset types, achieving the best performance in 5 out of 13 cases and the second-best results in 2 out of 8 cases. 
Moreover, we observe that $k$-Graph outperforms the baselines on average across the entire UCR-Archive for all measures. We also note that $k$-Graph has the best average rank for AMI and ARI and is second best for RI and ARI. Finally, $k$-Graph has the most significant number of wins (i.e., number of times $k$-Graph has the best performance) for ARI, AMI, and NMI, and the second largest number of wins for RI. Figure~\ref{fig:experiments_effectiveness_figures}(a) presents ARI, AMI, NMI, and RI across all datasets, demonstrating $k$-Graph's overall improvement compared to other approaches. Figure~\ref{fig:experiments_effectiveness_figures}(b) shows the critical difference diagram using a pairwise Wilcoxon sign rank test, with methods ranked along the horizontal axis.
Figure~\ref{fig:pairwise_comp} compares $k$-Graph's ARI with top-performing methods ($k$-Shape, SPF, and BIRCH). Each point represents a dataset's performance between two methods, with the green area indicating superior performance for y-axis methods and the white area for x-axis methods. Dotted lines mark 0.1 performance difference intervals, with points outside these bands indicating significant performance differences. The average distance from the identity line indicates the magnitude of performance differences between methods.

\begin{table*}[]
\scalebox{0.79}{
\begin{tabular}{r|c|c|c|c|c|c|c|c|c|c|c|c|c|c|c|c}
\hline
\textbf{}                             &\textbf{$k$-Graph}& $k$-Shape    & SPF            & BIRCH          & G.M.           & MB-$k$-M.   & $k$-Means   & SomTimes & T2F & DTC & OPTICS & Reservoir & Aggl. & HDBSC. & M.Sh. & DBSC. \\ \hline
\multicolumn{17}{c}{{\it Averaged ARI per category of the UCR-Archive datasets}} \\ \hline
\multicolumn{1}{r|}{\textbf{AUDIO}}   & 0.226          & 0.223          & 0.262          & \textbf{0.392} & {\ul 0.386}    & 0.329          & 0.343       & 0.074 & 0.118 & 0.320 & 0.000 & 0.000 & -0.000 & 0.012 & -0.000 & 0.000 \\
\multicolumn{1}{r|}{\textbf{DEVICE}}  & 0.101          & 0.072          & 0.080          & 0.063          & {\ul 0.108}    & 0.094          & 0.058       & \textbf{0.117} & 0.083 & 0.058 & 0.007 & 0.016 & 0.003 & 0.015 & 0.000 & 0.001 \\
\multicolumn{1}{r|}{\textbf{ECG}}     & {\ul 0.315}    & \textbf{0.363} & 0.248          & 0.285          & 0.279          & 0.243          & 0.251       & 0.224 & 0.166 & 0.190 & 0.112 & 0.082 & 0.051 & 0.011 & 0.000 & 0.000 \\
\multicolumn{1}{r|}{\textbf{EOG}}     & 0.129          & 0.152          & 0.126          & 0.156          & \textbf{0.188} & 0.163          & {\ul 0.170} & 0.146 & 0.139 & 0.139 & 0.000 & 0.065 & 0.000 & 0.021 & 0.000 & 0.000 \\
\multicolumn{1}{r|}{\textbf{EPG}}     & 0.445          & 0.035          & 0.416          & \textbf{1.000} & \textbf{1.000} & \textbf{1.000} & 0.182       & \textbf{1.000} & 0.538 & 0.109 & {\ul 0.772} & 0.759 & \textbf{1.000} & \textbf{1.000} & 0.000 & 0.000 \\
\multicolumn{1}{r|}{\textbf{H.DYN.}}  & \textbf{0.373} & 0.037          & {\ul 0.235}    & 0.109          & 0.092          & 0.087          & 0.043       & 0.066 & 0.048 & -0.000 & 0.001 & 0.000 & 0.018 & 0.005 & 0.000 & 0.000 \\
\multicolumn{1}{r|}{\textbf{IMAGE}}   & 0.290          & {\ul 0.297}    & 0.292          & \textbf{0.299} & 0.283          & 0.249          & 0.259       & 0.238 & 0.191 & 0.176 & 0.157 & 0.105 & 0.024 & 0.064 & 0.014 & -0.000 \\
\multicolumn{1}{r|}{\textbf{MOTION}}  & \textbf{0.170} & 0.154          & {\ul 0.156}    & 0.155          & 0.123          & 0.138          & 0.141       & 0.124 & 0.090 & 0.139 & 0.048 & 0.033 & 0.034 & 0.050 & 0.000 & 0.000 \\
\multicolumn{1}{r|}{\textbf{OTHER}}   & \textbf{0.945} & 0.414          & 0.308          & {\ul 0.713}    & 0.627          & 0.668          & 0.708       & 0.212 & 0.478 & 0.295 & 0.242 & 0.000 & 0.705 & 0.056 & 0.000 & 0.000 \\
\multicolumn{1}{r|}{\textbf{SENSOR}}  & \textbf{0.352} & 0.283          & {\ul 0.337}    & 0.224          & 0.238          & 0.215          & 0.234       & 0.226 & 0.191 & 0.174 & 0.166 & 0.074 & 0.097 & 0.093 & 0.001 & -0.005 \\
\multicolumn{1}{r|}{\textbf{SIM.}}    & {\ul 0.403}    & 0.353          & 0.376          & 0.315          & 0.351          & 0.274          & 0.261       & 0.292 & \textbf{0.418} & 0.209 & 0.110 & 0.113 & 0.070 & 0.034 & 0.000 & 0.000 \\
\multicolumn{1}{r|}{\textbf{SOUND}}   & \textbf{0.044} & 0.035          & {\ul 0.041}    & 0.004          & 0.011          & 0.010          & 0.016       & 0.031 & 0.028 & 0.000 & 0.000 & 0.016 & 0.001 & 0.000 & 0.000 & 0.000 \\
\multicolumn{1}{r|}{\textbf{SPECTRO}} & 0.208          & 0.167          & \textbf{0.234} & 0.133          & 0.130          & 0.172          & {\ul 0.212} & 0.170 & 0.171 & 0.063 & 0.157 & -0.001 & 0.020 & 0.054 & 0.000 & 0.000 \\
\multicolumn{1}{r|}{\textbf{TRAFFIC}} & 0.492          & 0.676          & 0.136          & 0.461          & {\ul 0.749}    & \textbf{0.757} & 0.047       & 0.260 & -0.018 & 0.054 & 0.000 & 0.000 & 0.008 & 0.098 & 0.008 & 0.000 \\ \hline
\multicolumn{17}{c}{{\it Averaged Accuracy for each dataset of UCR-Archive}} \\ \hline
\multicolumn{1}{r|}{\textbf{RI}}      & \textbf{0.729} & 0.702          & {\ul 0.721}    & 0.697          & 0.707          & 0.700          & 0.695       & 0.688 & 0.693 & 0.630 & 0.438 & 0.467 & 0.411 & 0.464 & 0.314 & 0.317 \\
\multicolumn{1}{r|}{\textbf{ARI}}     & \textbf{0.275} & 0.237          & {\ul 0.252}    & 0.238          & 0.239          & 0.223          & 0.208       & 0.209 & 0.181 & 0.145 & 0.124 & 0.075 & 0.061 & 0.069 & 0.004 & -0.001 \\
\multicolumn{1}{r|}{\textbf{AMI}}     & \textbf{0.315} & 0.291          & 0.291          & 0.296          & {\ul 0.293}    & 0.278          & 0.256       & 0.264 & 0.230 & 0.204 & 0.161 & 0.098 & 0.093 & 0.119 & 0.005 & 0.001 \\
\multicolumn{1}{r|}{\textbf{NMI}}     & \textbf{0.335} & 0.312          & 0.312          & {\ul 0.320}    & 0.317          & 0.302          & 0.282       & 0.284 & 0.259 & 0.217 & 0.165 & 0.102 & 0.118 & 0.133 & 0.006 & 0.001 \\ \hline
\multicolumn{17}{c}{{\it Number of win on the UCR-Archive}} \\ \hline
\multicolumn{1}{r|}{\textbf{RI}}      & {\ul 22} &	8  &	\textbf{26} &	8  &	4  &	4 &	6 &	5 &	8 &	4 &	2 &	1 &	1 &	2 &	0 &	1  \\
\multicolumn{1}{r|}{\textbf{ARI}}     & \textbf{25} &	11 &	{\ul 17} &	9  &	5  &	7 &	8 &	6 &	8 &	4 &	3 &	1 &	0 &	2 &	0 &	0 \\
\multicolumn{1}{r|}{\textbf{AMI}}     & \textbf{21} &	11 &	12 &	{\ul 13} &	12 &	1 &	7 &	4 &	7 &	3 &	7 &	4 &	0 &	3 &	1 &	0 \\
\multicolumn{1}{r|}{\textbf{NMI}}     & \textbf{21} &	12 &	12 &	{\ul 13} &	11 &	1 &	6 &	4 &	7 &	3 &	7 &	4 &	0 &	4 &	1 &	0 \\
\hline
\end{tabular}}
\caption{\rev{Effectiveness (ARI, RI, AMI, and NMI) per category of the UCR-Archive, the average (of measures) across all UCR-Archive, and the number of wins (i.e., the number of time methods achieved best accuracy) on the UCR-Archive. In bold, the best value per dataset type. The underlined represents the second best result.}}
\label{tab:effectiveness_category}
\vspace{-0.4cm}
\end{table*}

\etitle{Discussion} The results indicate an enhancement in accuracy with the implementation of $k$-Graph. 
This improvement is evident when evaluating performance across different dataset types and considering the overall average across all datasets. 
Nevertheless, we observe in Figure~\ref{fig:experiments_effectiveness_figures}(b) that, across all datasets of the UCR-Archive taken individually, $k$-Graph outperforms the baselines for NMI and AMI, albeit not significantly. For ARI and RI, SPF is slightly above $k$-Graph.
\paulJournal{It is also important to note that some of the traditional baselines demonstrate strong performances (across all measures) although these approaches were specifically proposed for time series. Nevertheless, among the top performing methods (ranked first or second in Figure~\ref{fig:experiments_effectiveness_figures}(a)), 3 approaches on 4 have been proposed for time series clustering.} \rev{Moreover, we observe that $k$-Graph performance is lower than the baselines for the TRAFFIC category. The latter is composed of one dataset (Chinatown) that contains very short time series (24 data points). As k-Graph relies on extracting meaningful subsequences, such dataset with very short time series is particularly hard to handle. Consequently, we can empirically observe a potential limitation of k-Graph for very short time series.}

Furthermore, the comparison illustrated in Figure~\ref{fig:pairwise_comp} with $k$-Shape provides intriguing insights. It unveils scenarios where $k$-Graph outperforms $k$-Shape, SPF, BIRCH, and vice versa, highlighting each approach's distinctive strengths and weaknesses when applied to diverse datasets. 
The latter emphasizes two key points: (i) the importance of an adaptable and robust approach like $k$-Graph that meets the objectives outlined in the initial challenge across various datasets, and (ii) the complementary nature of $k$-Graph, $k$-Shape, and BIRCH, suggesting potential synergies for improved accuracy.

\begin{figure}
    \centering
    \includegraphics[width=\linewidth]{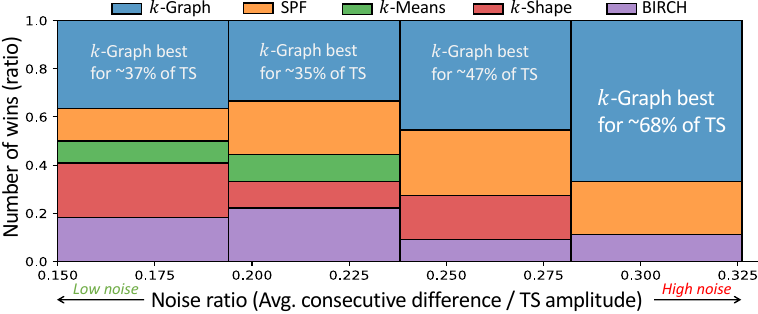}
    \caption{\rev{Number of win (ratio) versus Noise (ratio) in the UCR-Archive}}
    \vspace{-0.3cm}
    \label{fig:noise_exp}
\end{figure}

\rev{We investigate the performance differences between $k$-Graph and the four best-performing baselines by assessing the impact of noise on accuracy. The noise level of datasets in the UCR-Archive is measured by the average noise ratio, defined as the ratio of the average difference between consecutive points to the maximum amplitude of the time series. We calculate the number of wins (i.e., instances where a method achieves the best ARI score) for datasets within predefined noise ratio intervals. As shown in Figure~\ref{fig:noise_exp}, the proportion of $k$-Graph wins increases with the noise ratio, reaching 68\% for time series with a noise ratio above 0.28. Overall, Figure~\ref{fig:noise_exp} indicates that $k$-Graph's efficiency is unaffected by noise and is more robust than the baselines.}

\subsection{Execution Time Evaluation}

\begin{figure}[tb]
 \centering
\includegraphics[width=0.9\linewidth]{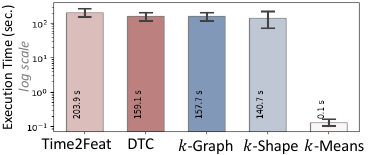}
\vspace{-0.2cm}
 \caption{Execution time computed on all datasets of the UCR-Archive.}
\vspace{-0.3cm}
 \label{fig:exec_time}
\end{figure}

\etitle{Presentation} We compare our method's execution time against other approaches, all configured for a multi-core CPU setup. To ensure consistency, we focus on methods with pure Python implementations, including Time2Feat, $k$-Shape, and DTC, \rev{but also include $k$-Means (implemented in C) for a comprehensive evaluation.} Figure~\ref{fig:exec_time} shows that $k$-Graph, $k$-Shape, and DTC have similar execution times, differing by about 10 seconds. Time2Feat is the slowest, taking 50 seconds longer, \rev{while $k$-Means is the fastest, likely due to its C implementation.}
\etitle{Discussion} The observed results indicate that extracting numerous graphs does not compromise the algorithm's efficiency, as it maintains competitive performance compared to widely adopted methods, \dt{as required by the first challenge. However, it is noteworthy that $k$-Means, employing a straightforward solution, outperforms other approaches regarding speed.}

\subsection{Parameter Influence}

\begin{figure}[tb]
 \centering
\includegraphics[width=\linewidth]{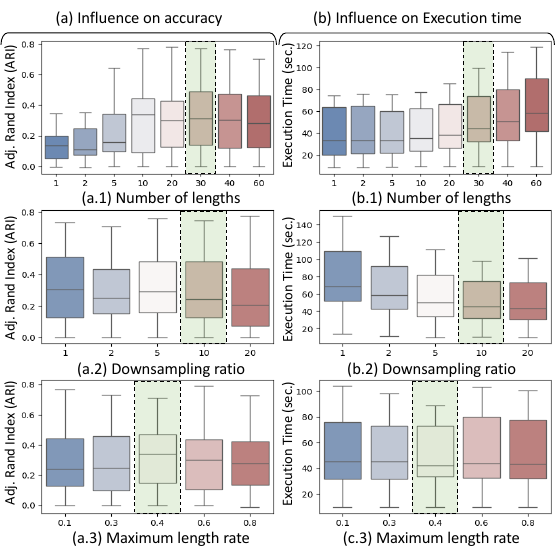}

 \caption{Influence of $k$-Graph parameters on (a) accuracy and (b) execution time (green hover the default values of these parameters).}
 \label{fig:param_infl}
\vspace{-0.3cm}
\end{figure}

\etitle{Presentation} As introduced in Section \ref{sec:proposed}, $k$-Graph has three main parameters: Number of Lengths ($M$) for randomly selected subsequence lengths, Sample ($smpl$) to reduce subsequences in PCA training, and Rate Maximum Length ($rml$) for the upper limit of subsequence length selection. Figure \ref{fig:param_infl} demonstrates parameter impacts across UCR Datasets, showing effectiveness (ARI) in column (a) and efficiency (seconds) in column (b). The rows show variations in $M$ (eight values), $smpl$ (five values), and $rml$ (five values) parameters respectively. For consistency, default values of $M=30$, $smpl=10$, and $rml=0.4$ are maintained when not studied.

\begin{figure}[tb]
 \centering
\includegraphics[width=\linewidth]{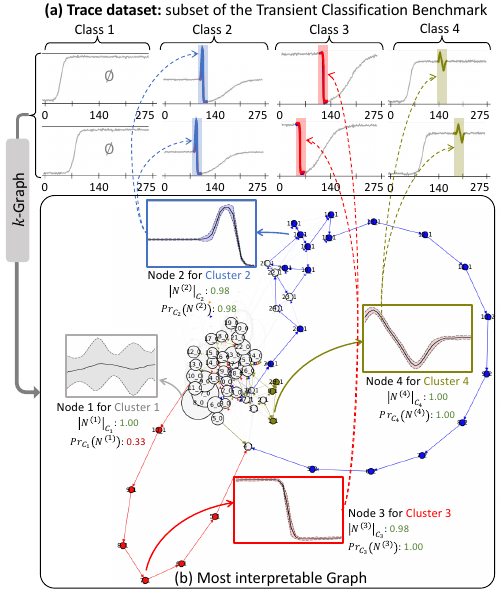}

 \caption{$k$-Graph interpretability for the Trace dataset from the UCR-Archive. The highlighted nodes are the most representative and exclusive subsequences identified by $k$-Graph for each cluster.}
\label{fig:interpretability}
\vspace{-0.3cm}
\end{figure}

\etitle{Discussion} The experimentation with the parameter M reveals that increasing the number of lengths positively impacts accuracy until it reaches a plateau at around $M=30$ lengths. Beyond this point, the accuracy shows little improvement. Correspondingly, as the lengths increase, the time required by the algorithm also increases, demonstrating a trade-off between accuracy and computational efficiency. A favorable balance between execution time and performance is achieved with approximately 30 different lengths. Examining the $smpl$ parameter indicates that reducing the number of subsequences minimally impacts performance until around 10. However, it significantly affects the algorithm's execution time. As we decrease the number of subsequences, the time performance improves. Achieving a good balance involves reducing the number of subsequences by 10 times. Finally, the analysis of the $rml$ parameter shows that adopting 40\%(0.4) of the length of the shortest series in the dataset as the maximum subsequence length is the optimal solution. This choice provides similar execution times while demonstrating superior performance in terms of effectiveness.

\subsection{$k$-Graph Interpretability}

\etitle{Presentation} In this section, we explore how $k$-Graph enhances the interpretability of clustering results compared to $k$-Shape. While $k$-Shape and $k$-Graph differ in interpretability---$k$-Graph offers a graph-based, subsequence-based interpretation, whereas $k$-Shape provides centroids of the same size as the dataset's time series---we still present $k$-Shape's centroids to highlight the interpretability advantage of $k$-Graph. Figure~\ref{fig:interpretability} provides visual insights using the Trace dataset from the UCR-Archive, which contains simulated time series representing various instrument failures in nuclear power plants (four classes). 
In this figure, we observe the graph $\mathcal{G}_{\bar{\ell}}$ for the optimal length $\bar{\ell}=36$, along with the most representative and exclusive nodes for each cluster ($\bar{N}_{C_1}=N^{(1)}$, $\bar{N}_{C_2}=N^{(2)}$, $\bar{N}_{C_3}=N^{(3)}$, $\bar{N}_{C_4}=N^{(4)}$).
The colored nodes in Figure~\ref{fig:interpretability}  represent those with an exclusivity above 0.9 (i.e., the $\gamma$-graphoids with $\gamma=0.9$). 

\begin{figure*}[tb]
 \centering
\includegraphics[width=\linewidth]{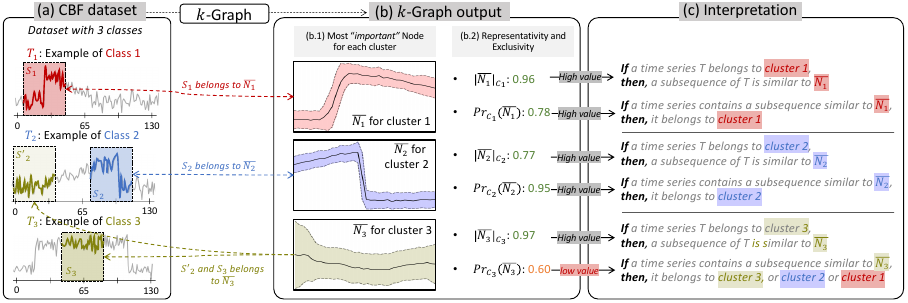}
\vspace{-0.3cm}
 \caption{\paulJournal{Interpretability of $k$-Graph clustering applied to the CBF dataset.}}
\vspace{-0.3cm}
\label{fig:interpretability_practice}
\end{figure*}

\begin{figure}[tb]
 \centering
\includegraphics[width=\linewidth]{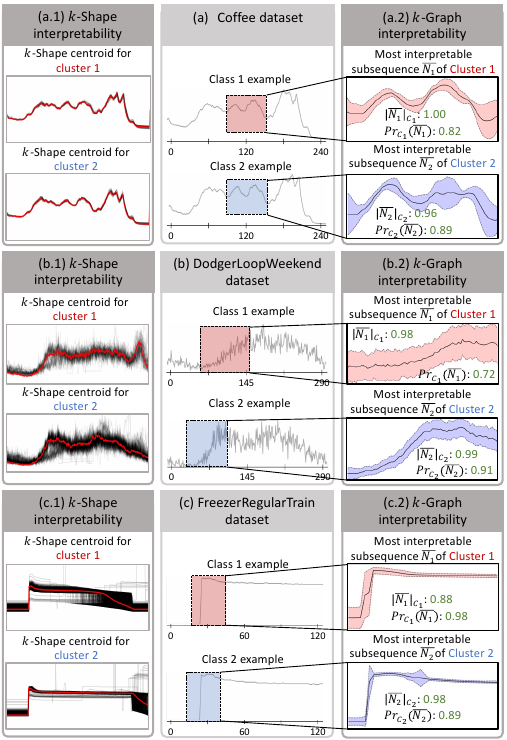}
\vspace{-0.4cm}
 \caption{\rev{Comparisons of $k$-Graph and $k$-Shape interpretability for each cluster obtained on three UCR-Archive datasets. (*.1) centroids (in red) computed using $k$-Shape.  (*.2) most representative and exclusive node $\bar{\mathcal{N}}$.}}
\vspace{-0.5cm}
\label{fig:interpretability_ex}
\end{figure}

\etitle{Discussion} In Figure~\ref{fig:interpretability}, most nodes represent subsequences from all clusters (i.e., grey nodes), but some, like Node $ N^{(2)} $, show high representativity and exclusivity. This figure highlights $k$-Graph's interpretability capabilities, showing that its clustering partition aligns with the labels of the Trace dataset. The graphoids effectively identify the discriminant features of clusters 2, 3, and 4. Notably, Node $ N^{(4)} $ has $ |N^{(4)}|_{C_4}=1 $ and $ Pr_{C_4}(N^{(4)})=1 $, indicating it is crossed only by time series from cluster 1.

For cluster 1, the distinctive characteristic is the absence of discriminative patterns, as indicated by Node $ N^{(1)} $ with $ |N^{(1)}|_{C_1}=1 $ and $ Pr_{C_1}(N^{(1)})=0.33 $. While this node is highly representative, its low exclusivity suggests that cluster 1 lacks specific patterns, which aligns with the unique nature of class 1 in the Trace dataset. This ability to detect clusters without discriminant subsequences highlights $k$-Graph's robustness in handling diverse data structures. The representativity and exclusivity measures are crucial for users to evaluate the validity of identified nodes, providing a mechanism to filter out potential misinterpretations and ensuring that the clustering results are both meaningful and actionable.

\subsection{Interpretability in Practice}

\paulJournal{
Even though the graph can be important for the user to interact and explore the clustering result, the graph is not necessary for interpreting the clustering partition. In practice, we only need to return the most ``important" nodes (i.e., with the highest representativity and exclusivity).}

\paulJournal{
Figure~\ref{fig:interpretability_practice} depicts the $k$-Graph output on the CBF datasets (illustrated in Figure~\ref{fig:interpretability_practice}(a)).
$k$-Graph can automatically select the most representative and exclusive node for each cluster (illustrated in Figure~\ref{fig:interpretability_practice}(b)).
Each node is associated with its corresponding exclusivity ($Pr_{C_i}(N)$) and representativity ($|N|_{C_i}$) values.
These values give a direct interpretation of the clustering. For example, the high representativity ($|\bar{N_1}|_{C_1}=0.96$) and high exclusivity ($Pr_{C_1}(\bar{N_1})=0.78$) of $\bar{N_1}$ imply the following interpretation: {\it "A time series belongs to cluster 1 if and only if a subsequence of this time series is similar to $\bar{N_1}$"}. We can see indeed in Figure~\ref{fig:interpretability_practice}(a) that a subsequence of $T_1$ (called $S_1$) is similar to $\bar{N_1}$. 

On the contrary, the high representativity ($|\bar{N_3}|_{C_3}=0.96$) and medium exclusivity ($Pr_{C_3}(\bar{N_3})=0.60$) of $\bar{N_3}$ imply the following interpretation: {\it "A time series belongs to cluster 3 if a subsequence of this time series is similar to $\bar{N_3}$. However, a time series $T$ containing a subsequence similar to similar to $\bar{N_3}$ does not necessarily mean that T belongs to cluster 3"}. Figure~\ref{fig:interpretability_practice}(a) confirms that a subsequence of $T_2$ and $T_3$ (called $S'_2$ and $S_3$ respectively) are similar to $\bar{N_3}$. However, $T_2$ belongs to cluster 2, and $T_3$ belongs to cluster 3. 

In general, $k$-Graph helps us to understand that the CBF dataset is composed of (i) time series with a rapid increase followed by a slow decrease (class 1 matched by cluster 1), (ii) time series with a slow increase followed by a rapid decrease (class 2 matched by cluster 2), and (iii) time series that do not contains rapid increase or decrease, and that contain a flat subsequence (class 3 matched by cluster 3).
}

\subsection{Interpretability Examples}

\etitle{Presentation}
\paul{In Figure~\ref{fig:interpretability_ex}, we provide comparisons between $k$-Shape's and $k$-Graph's interpretability on three additional datasets of the UCR-Archive: (a) Coffee dataset with two classes, (b) DodgerLoopWeekend datasets with two classes, (c) 
FreezerRegularTrain datasets with two classes.
For $k$-Shape, interpretability is driven by inspecting the centroids of each cluster.
For $k$-Graph, we provide the most representative and exclusive nodes (due to lack of space, we omit plotting the corresponding graphs) for each cluster obtained with $k$-Graph.}

\etitle{Discussion} Figure~\ref{fig:interpretability_ex} highlights the most representative nodes for each cluster in four UCR-Archive datasets. In all cases, $k$-Graph provides clustering consistent with the labels. For DodgerLoopWeekend
(Figure~\ref{fig:interpretability_ex}(b.2)),
the nodes clearly correspond to class-specific patterns. 

In contrast, the Coffee and FreezerRegularTrain datasets (Figure~\ref{fig:interpretability_ex}(a) and (c)) present more subtle patterns. In the Coffee dataset, the nodes reveal that class differences lie in the relative heights of two central peaks. 

For FreezerRegularTrain, despite similar time series, $k$-Graph identifies nodes that highlight subtle differences in the initial increase, with cluster 2 showing a more abrupt rise and longer plateau. These nodes are highly representative and exclusive for their clusters ($|\bar{N_1}|_{C_1}=0.88$, $|\bar{N_2}|_{C_2}=0.98$ and $Pr_{C_1}(\bar{N_1})=0.98$, $Pr_{C_2}(\bar{N_2})=0.89$), aligning with discriminative features noted in the UCR-Archive~\cite{Dau2018TheUT}. These examples demonstrate the effectiveness of $k$-Graph in identifying relevant patterns for complex tasks.

\paul{Unlike $k$-Graph, the centroids computed with $k$-Shape often lack sufficient information to distinguish between classes. 
In examples, such as Coffee, DodgerLoopWeekend, and FreezerRegularTrain, distinguishing between classes is challenging. Additional comparisons on UCR-Archive datasets are available in our GitHub repository. This analysis suggests that $k$-Graph offers improved interpretability over previous solutions.}

%% file: Conclusion.tex
\section{Conclusions}
\label{sec:conclusions}

Despite significant interest in time series clustering, many existing methods lack interpretability. We introduce $k$-Graph, a novel graph-based approach that clusters time series datasets while providing interpretable clustering partitions. Our experiments show that $k$-Graph effectively addresses the challenges outlined in Section~\ref{sec:intro}. It combines the strengths of existing methods by preserving temporal dependencies and achieving performance on par with raw and deep learning approaches. Additionally, it enhances interpretability without sacrificing granularity, a common issue with feature-based methods. 

\rev{In future work, we aim to explore the use of Graph Neural Networks in our clustering process and extend our approach to multivariate time series.}
We note that applying $k$-Graph to multivariate time series is not straight-forward. 
The main challenge in doing so is due to the inter-dependencies among dimensions, which should be considered in the graph embedding step.
This would lead to a potential explosion of the graph size. 
We consider this problem non-trivial, and a very interesting future research direction.